\documentclass[]{bytedance_seed}
\usepackage{tabularx}

\usepackage[toc,page,header]{appendix}

\usepackage{minitoc}
\usepackage{cleveref} 
\usepackage{subcaption}
\usepackage{booktabs}
\usepackage{graphicx}
\usepackage{pgfplots}
\usepackage{pgfplotstable}
\usepackage{xcolor}
\usepackage{CJKutf8}
\usetikzlibrary{patterns}
\usepackage{float}
\usepackage{caption}
\usepackage{algorithm}
\usepackage{algpseudocode}
\newcommand{\method}{Astra\xspace}

\usepackage{natbib}
\usepackage{latexsym}

\usepackage{url}
\usepackage{amssymb}
\usepackage[utf8]{inputenc}
\usepackage{microtype}
\usepackage{booktabs}
\usepackage{pifont} 
\usepackage{multirow}
\usepackage{makecell}
\usepackage{paralist}
\usepackage{xspace}
\usepackage{color}
\usepackage{xcolor}
\usepackage{colortbl}
\usepackage{adjustbox}
\usepackage{hyperref} 
\usepackage[edges]{forest}
\usepackage{tikz} 
\usepackage{caption}
\usepackage{amsfonts}

\hypersetup{
    colorlinks,
    linkcolor={blue!80!black},
    citecolor={blue!80!black},
}
\tikzset{
    root/.style =             {align=center, text width=1cm, rounded corners=3pt, line width=0.3mm, fill=gray!10, draw=gray!80, font=\small},
    demographic/.style =         {align=center, text width=1.8cm, rounded corners=3pt, line width=0.3mm, fill=blue!10, draw=blue!80, font=\footnotesize},
    demographic_work/.style =    {align=center, text width=10cm, rounded corners=3pt, line width=0.3mm, fill=blue!10, draw=blue!0, font=\footnotesize},
    character/.style =         {align=center, text width=1.8cm, rounded corners=3pt, line width=0.3mm, fill=red!10, draw=red!80, font=\footnotesize},
    character_work/.style =    {align=center, text width=10cm, rounded corners=3pt, line width=0.3mm, fill=red!10, draw=red!0, font=\footnotesize},
    personalization/.style =           {align=center, text width=1.8cm, rounded corners=3pt, line width=0.3mm, fill=cyan!10, draw=cyan!80, font=\footnotesize},
    personalization_work/.style =      {align=center, text width=10cm, rounded corners=3pt, line width=0.3mm, fill=cyan!10, draw=cyan!0, font=\footnotesize},
    risk/.style =         {align=center, text width=1.8cm, rounded corners=3pt, line width=0.3mm, fill=orange!10, draw=orange!80, font=\footnotesize},
    risk_work/.style =    {align=center, text width=10cm, rounded corners=3pt, line width=0.3mm, fill=orange!10, draw=orange!0, font=\footnotesize},
}

\newcommand{\ie}{\textit{i.e.}\xspace}
\newcommand{\eg}{\textit{e.g.}\xspace}

\usepackage{CJK}

\newtcolorbox{conclusionbox}[1][]{
    colback=blue!5!white,    
    colframe=blue!50!black,  
    fonttitle=\bfseries,
    title=Conclusion,        
    enhanced,
    attach boxed title to top left={xshift=5mm, yshift=-2mm},
    boxed title style={colback=blue!30!white},
    #1 
}

\newcounter{conclusion}
\newenvironment{numberedconclusion}[1][]{
    \refstepcounter{conclusion}
    \begin{conclusionbox}[title=Main Results \theconclusion, #1]
}{
    \end{conclusionbox}
}
\newcommand{\imgcaseheight}{2cm}
\title{\method: Toward General-Purpose Mobile Robots via Hierarchical Multimodal Learning }

\affiliation{ByteDance Seed}

\contribution{Full author list in Contributions}

\abstract{

Modern robot navigation systems encounter difficulties in diverse and complex indoor environments. Traditional approaches rely on multiple modules with small models or rule-based systems and thus lack adaptability to new environments. To address this, we developed \method, a comprehensive dual-model architecture, \method-Global and \method-Local, for mobile robot navigation. \method-Global, a multimodal LLM, processes vision and language inputs to perform self and goal localization using a hybrid topological-semantic graph as the global map, and outperforms traditional visual place recognition methods.
 \method-Local, a multitask network, handles local path planning and odometry estimation. Its 4D spatial-temporal encoder, trained through self-supervised learning, generates robust 4D features for downstream tasks. The planning head utilizes flow matching and a novel masked ESDF loss to minimize collision risks for generating local trajectories, and the odometry head integrates multi-sensor inputs via a transformer encoder to predict the relative pose of the robot. Deployed on real in-house mobile robots, \method achieves high end-to-end mission success rate across diverse indoor environments.

\date{June 6, 2025}
\checkdata[Project Page]{\url{https://astra-mobility.github.io/}}

}

\begin{document}

\maketitle


\section{Introduction}

Modern robotic systems face increasing demands for adaptive navigation in diverse and complex environments. In a known environment, navigation can be decomposed into three major challenges. \textbf{Goal Localization}: in certain applications, instead of directly providing a pose for navigation goal, the goal may be specified by natural language or goal image prompt. In these cases, we need a system to understand the prompt and localize the goal within the map. \textbf{Self-Localization}: the robot needs to localize itself in the map. In complex scenarios like warehouses, the environment is highly repetitive with few global landmarks. Traditional navigation system often needs to rely on artificial landmarks like QR code. \textbf{Path Planning}: path planning can be divided into global planning and local planning. Global planning generates a coarse route based on the robot pose and goal pose. Given way points along global path, local path planning is responsible for reaching intermediate way points while avoiding obstacles.

To address goal localization, self-localization, and path planning tasks, traditional navigation systems generally consist of multiple modules, e.g., localization, perception, prediction, planning, and control. Different modules often contain multiple small models or rule-based systems. In recent years, the emergence of foundation models has spurred a trend to integrate small models into larger ones to solve more tasks. However, the question of how many models are necessary remains unanswered. In this report, we propose \method, a dual-model architecture that tackles the above three navigation problems in diverse indoor environments. \method adheres to the System 1/System 2 philosophy \cite{fastslow}, where \method-Global is responsible for low frequency tasks such as goal and self-localization, and \method-Local manages high frequency tasks including local path planning and odometry estimation.

\begin{figure}[t]
\centering
\small
\includegraphics[width=1\linewidth]{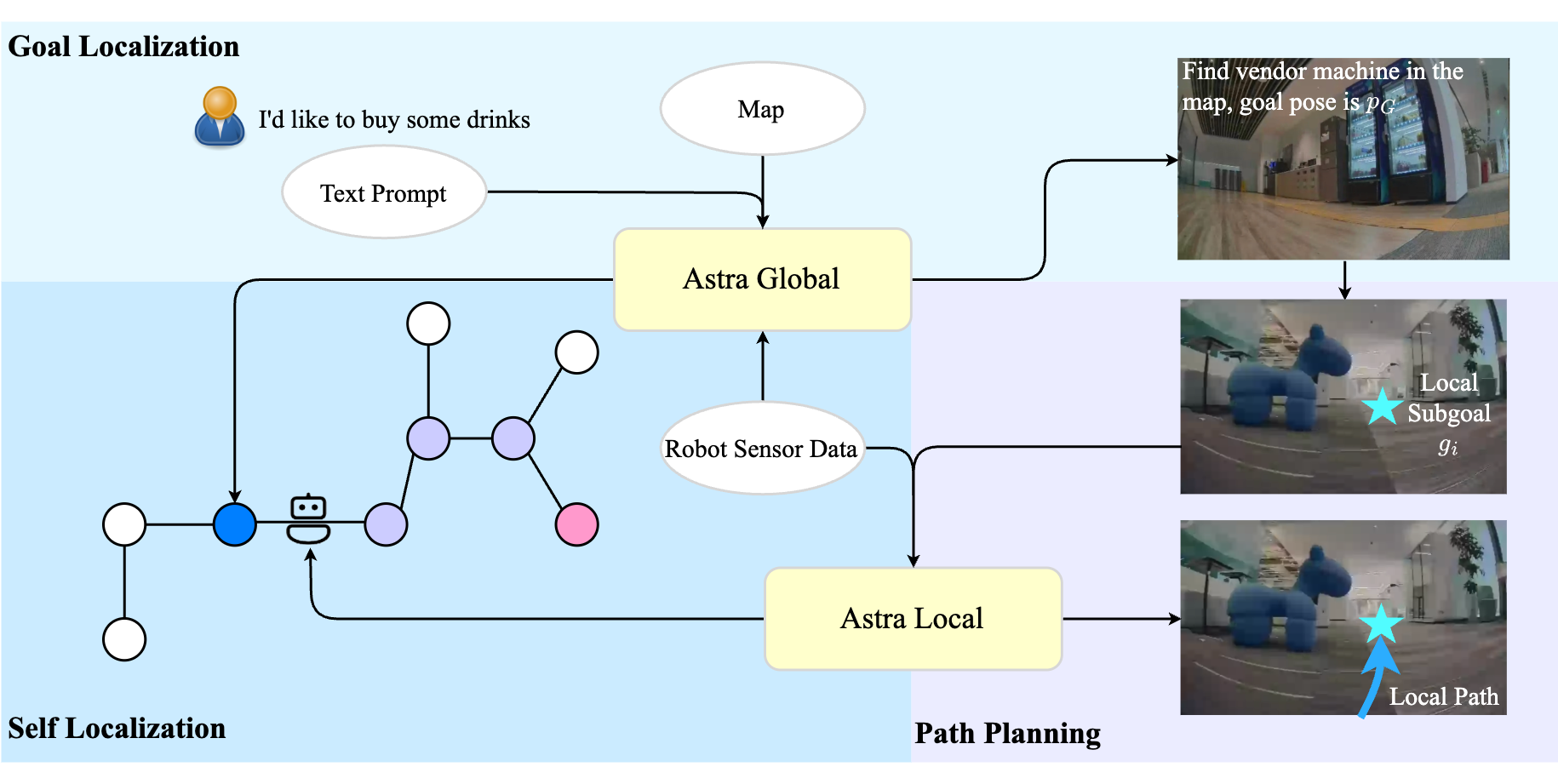}
\vspace{-12pt}
\caption{\method addresses three key navigation challenges: goal localization, self-localization, and path planning, with the holistic integration of \method-Global and \method-Local. \method-Global is responsible for goal and self-localization. For the goal localization, it locates the landmark and the corresponding goal pose $p_G$ from the map based on user text prompts. For self-localization, \method-Global identifies visual landmarks from images and then integrates this information with the odometry estimated by \method-Local through multi-sensor fusion to obtain the robot's global pose $p_i$. Meanwhile, \method-Local takes an additional subgoal $g_i$ as input for path planning and generates a local path for the robot to follow. }
\label{fig:astra_framework}
\end{figure}

\method-Global is a multimodal LLM (MLLM) responsible for self and goal localization within a global map. We represent the known environment as a topological-semantic map and use it directly as context input for the MLLM. Given this map representation along with a query image or text prompt, the model locates the query within the map. Whether the query is a text prompt from a user, such as ‘I'd like to find somewhere to rest’, in which case the model locates the goal in the map, or an image the robot currently perceives, in which case it locates the robot itself, \method-Global can handle both cases. Unlike traditional mobile robot global localization methods that often rely on artificial landmarks like QR codes or additional sensors in complex scenes, \method-Global leverages natural human built landmarks and functions effectively in diverse environments.

On the other hand, \method-Local is a multi-task network tasked for local path planning and odometry estimation. The model takes sensor inputs (e.g., multi-view images, IMU, wheel), a local goal, and other robot states as inputs. Based on these, it plans a local path and estimates the robot's odometry. \method-Local features a 4D spatial-temporal encoder that fuses images from a multi-camera setup across multiple frames. Pretrained on a large scale datasets, the encoder is capable of generating a coherent 4D spatial-temporal representation and predicting future representations. The encoder is connected to two task heads. The planning head, taking additional inputs such as the local goal and robot states, employs flow matching to generate a local path that guides the robot towards the local goal while avoiding obstacles. The odometry head uses a transformer to fuse vision features with other sensor inputs to estimate robot relative movement.

Our key contributions include:
\begin{itemize}
    \item A novel dual-model system \method that effectively solves key mobile navigation problems, as depicted in Fig.~\ref{fig:astra_framework}. We developed and deployed \method to our in-house built mobile robots and tested across different indoor environments including warehouse, office building, house. In all environments, \method achieves high end-to-end mission success rate from locating the user query to navigating to the goal safely. 
    \item \method-Global, a MLLM designed to handle low frequency tasks in mobile navigation, effectively addresses both goal and self-localization within a single model. Built upon a foundation MLLM, we train \method-Global using Supervised Finetuning (SFT) and Reinforcement Learning (RL). \method-Global outperforms traditional Visual Place Recognition (VPR) methods across all environments and works zero-shot in new environments. Our experiments demonstrate that RL is an effective approach for enhancing the model's generalization capabilities and is more data efficient than using SFT alone.
    \item \method-Local is a multi-task network designed for high frequency tasks in mobile navigation, such as odometry estimation and local path planning. Its shared encoder, pretrained on large scale datasets in a self-supervised manner, provides robust spatial-temporal visual features and outperforms state-of-the-art (SOTA) methods in downstream tasks like occupancy forecasting. The odometry head, which fuses camera, IMU, wheel data within a novel transformer framework, has been shown by experimental results to be effective in multi-sensor data fusion. The planning head employs flow matching for local path planning. When trained with a novel masked ESDF loss function, it achieves a significant reduction in the collision rate.
\end{itemize}

The remainder of this report is organized as follows: Section \ref{approach} elaborates on the details of \method, including the overall architecture, model design, and training procedures for both \method-Global and \method-Local. Section \ref{exp} presents the experimental results for the entire system and each model. Section \ref{related_work} reviews related studies, and we conclude in Section \ref{conclusion}.
\section{Related Work}
\label{related_work}

\subsection{Global Localization}
The goal of visual-based global localization, a critical component of visual navigation, is to determine the position of the current image within a known scene. Traditional approaches like Visual Place Recognition \cite{ali2023mixvpr, wang2022transvpr, hausler2021patch} rely on image retrieval to match the current scene with the most similar image in a pre-built map, offering robustness to environmental and seasonal changes but suffering from limitations such as loss of image details in global descriptions—leading to poor performance in repetitive scenes with minor differences—and an inability to directly output camera pose as they focus on retrieval rather than pose estimation. In contrast, end-to-end (E2E) methods have emerged to directly estimate ego pose from sensor inputs, leveraging architectures that bypass complex geometric calculations (e.g., PixLoc \cite{lindenberger2021pixel}, BEV-Locator \cite{zhang2025bev}, EgoVM \cite{he2024egovm}, MapLocNet \cite{wu2024maplocnet}). While these E2E approaches streamline the pipeline and enable direct pose output, they often struggle with generalization and performance degradation under significant scene and season variations, highlighting the need for more robust strategies to balance accuracy and adaptability across diverse environments.

Recently, the rapid advancement of Vision-Language Models (VLMs) \cite{bai2025qwen2, team2024gemini, steiner2024paligemma}, powered by large-scale pretraining on billions of image-text pairs, has revolutionized cross-modal understanding in robotics domain. While existing VLM research focuses on navigation \cite{zhou2024navgpt, yin2025unigoal} and manipulation \cite{kim2024openvla,qu2025spatialvla}, their application to visual localization remains underexplored. Works like MobilityVLA \cite{mobilityvla} use long-context VLMs for topological localization but struggle with low precision due to semantic-geometric mismatches. Our \method-Global addresses this by introducing a two-stage localization framework paired with a rigorously constructed multi-scenario dataset, enhancing both accuracy and real-world adaptability.

\subsection{Odometry Estimation}
Traditional multi-sensor fusion odometry methods typically rely on probabilistic frameworks like Kalman filters, particle filters, or factor graph optimization to integrate data from LiDAR, cameras, and IMU, etc. Existing multi-sensor fusion odometry frameworks, such as filtering-based approaches exemplified by R3LIVE \cite{lin2022r} and factor graph optimization methods like LVI-SAM \cite{shan2021lvi}, demonstrate satisfactory accuracy under nominal operating conditions. However, these methodologies predominantly rely on handcrafted feature extraction and incorporate simplified assumptions in uncertainty modeling. Such limitations hinder their capacity to fully characterize the intrinsic sensor observation properties, leading to performance degradation in edge cases such as geometrically degenerate environments or adverse weather conditions.

Recent advancements in deep learning have catalyzed paradigm shifts in state estimation. While single-modality odometry solutions (e.g., DROID-SLAM \cite{teed2021droid}, DPVO \cite{teed2023deep}, BEV-ODOM \cite{bevodom}, RoNIN \cite{herath2020ronin}) demonstrate domain-specific competence, they exhibit limitations in accuracy and robustness compared to multi-sensor counterparts which effectively exploit complementary sensor characteristics. Transformer-based architectures, in particular, have gained prominence due to their superior cross-temporal correlation modeling and cross-modal interaction capabilities, such as TransFusionOdom \cite{sun2023transfusionodom} and VIFT \cite{kurt2024causal}. The integration of temporal dynamics with multi-modal fusion presents a promising direction for advancing odometry systems. Therefore, our method integrated extended multi-view and multi-modal temporal data to achieve superior accuracy and robustness.

\subsection{End-to-End Planning}

End-to-end planning offers a compelling vision for mobile robot navigation, as well as autonomous driving and robot manipulation, by simpler and more adaptive systems.
From early demonstrations \cite{BojarskiTDFFGJM16} to recent innovations incorporating transformers, diffusion models and LLMs \cite{HuCVPR23, zhao2023learning, black2024pi_0, XuRAL24}, the field has made significant strides. 
Industry efforts, such as Tesla’s Full Self-Driving (FSD) \cite{fsd}, leverage end-to-end learning to process vast datasets, aiming for scalable performance \cite{LanCoRR24}.
However, since most approaches apply imitation learning that relies on massive expert data, challenges like out-of-distribution problems, system robustness and interpretability of results remain, requiring ongoing research to ensure practical deployment \cite{ChenPAMI24}. 
Comparing with existing methods, our proposed masked ESDF loss, can significantly reduce collision rates while preserving high modality of trajectories.

\subsection{3D and 4D Encoders}
In vision-based 3D occupancy prediction, camera inputs are utilized to predict the occupancy status, offering a cost-effective alternative to LiDAR-based systems \cite{cao2022monoscene, huang2023tri, wei2023surroundocc}. Recently, self-supervised learning for occupancy gained widespread attention. Self-supervised learning addresses the data scarcity challenge by enabling models to learn from unlabeled data, reducing the need for costly manual annotations. The volume rendering technique is generally adopted, which back-projects the 3D voxel volume into the 2D image space (represented by depth) \cite{yang2024unipad, pan2024renderocc}. Then, the rendered depth can be optimized using photometric consistency or temporal coherence \cite{huang2024selfocc, zhang2023occnerf}. 

4D occupancy forecasting extends 3D occupancy prediction into the temporal domain, predicting how the environment will evolve over time. In \cite{zheng2024occworld},  a GPT-like spatial-temporal generative transformer is utilized to generate subsequent scene and ego tokens, which are decoded into the future occupancy and ego trajectory simultaneously. \cite{yang2025driving} predicts future occupancy and flow, conditioned on ego-vehicle actions like velocity and steering angle, and integrates with end-to-end planning. In \cite{ma2024cam4docc}, a benchmark for camera-only 4D occupancy forecasting is proposed. It evaluates sequential occupancy states and 3D backward centripetal flow, highlighting challenges in long-term prediction. Different from these methods, our proposed 4D Spatial-Temporal encoder in Astra-Local does not require 3D semantic labels, which is a highly cost-effective solution.
\section{Approach}
\label{approach}

We now formally formulate the mobile navigation problem. We consider a wheeled mobile robot operating in a pre-mapped environment, for which one or more demonstration tour videos are available. Given an instruction \textit{I} in the form of a text prompt from the user, the robot's task is two-fold. First, it needs to locate the goal. Then, it must navigate towards the goal while avoiding any obstacles in its path.

Fig.~\ref{fig:astra_framework} illustrates the overall navigation process with \method. Given a demonstration tour of the indoor environment, we first build a map \textit{G} offline. When a user provides a prompt, \method-Global searches the map \textit{G} for landmarks that match the prompt and sets the corresponding pose as the navigation goal. For self-localization, \method-Global performs low frequency global localization of the robot in the map using robot's onboard sensor data, specifically images. Meanwhile, \method-Local fuses images with other sensors such as IMU, wheel to predict the local relative pose at a high frequency. By combining these two, we obtain high frequency localization results. This mirrors human behavior in most environments, where we use sparse landmarks for global positioning and estimate our current location through dead-reckoning, calculating relative movements (odometry) from our last known position. In the path planning task, a simple search-based global path planner generates a route to the goal and selects a local subgoal based on the robot's current pose. \method-Local then uses sensor data to generate a local path to the subgoal while avoiding obstacles. Algorithm \ref{alg:astra} shows the main navigation loop.

\begin{algorithm}
\caption{\method for mobile navigation}
\label{alg:astra}
\begin{algorithmic}
\Require Map \(\mathit{G}\), user instruction \(\mathit{I}\), robot sensor data \(O\)

\noindent Locate goal pose \(\mathbf{p_G}\) from \(\mathbf{I}\) and robot pose \(\mathbf{p_0}\) with \method Global

\noindent Plan a global trajectory \(\mathbf{Tr}\)
\While{\(\mathbf{p_G}\) is not reached}

Select a subgoal \(\mathbf{g_i}\) from \textbf{Tr} based on \(\mathbf{p_{i-1}}\)

Plan a local trajectory to the \(\mathbf{g_i}\) using \method-Local

Update \(\mathbf{p_i}\) with \method-Local and \method-Global

\EndWhile
\end{algorithmic}
\end{algorithm}

\subsection{\method-Global}
\label{sec:astra_global}
\begin{figure}[!ht]
\centering
\includegraphics[width=0.8\textwidth]{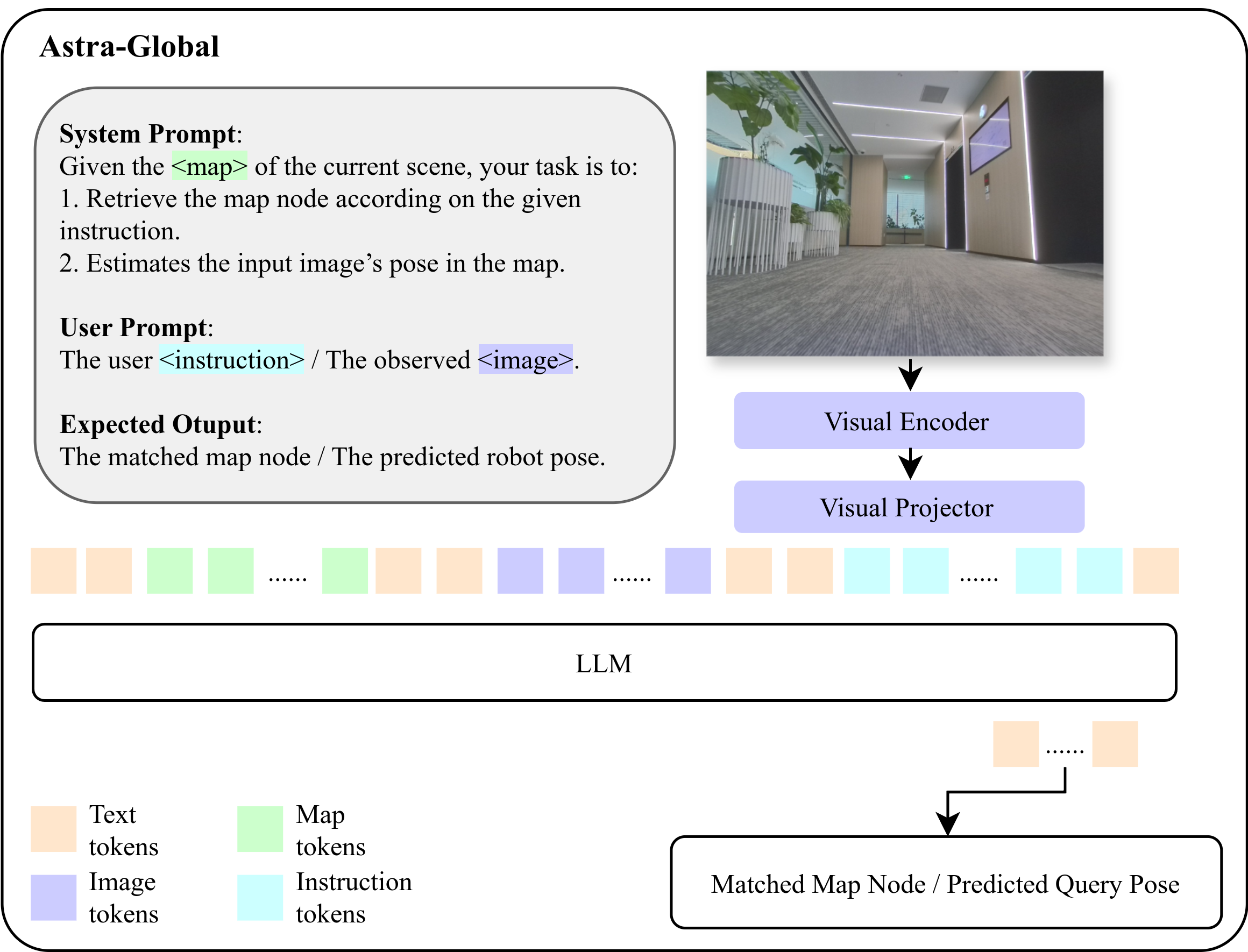}
\caption{\method-Global follows most modern MLLMs like \cite{bai2025qwen2} where images are encoded via a separate visual encoder and further aligned with text tokens with a projector. Map is represented as a combination of images and texts depending on localization stage. The encoded vision tokens and text tokens are fed into a LLM to generate the final results. }
\label{fig:astra_global}
\end{figure}

\method-Global focuses on two tasks: self-localization and goal localization. Traditional indoor localization methods either depend on artificial landmarks such as QR codes or require a complex map-building process, making the entire system difficult to deploy or adapt to new environments. In contrast, humans often localize themselves in complex scenes using high-level semantic landmarks. Inspired by this and the fact that self and goal localization only need to operate at a low frequency, we design \method-Global as a MLLM as shown in Fig.~\ref{fig:astra_global}. Thanks to recent advancements, MLLMs have excellent scene understanding and grounding capabilities and can handle multimodal inputs naturally. We first create a comprehensive map that includes geometric poses, visual landmarks, and connectivity constraints. This map provides the foundation for the robot to understand the scene. By using the map as context input for \method-Global, it enables vision-language localization for prior-free self-localization and language-based goal localization. Next, we introduce the offline mapping module.

\subsubsection{Offline Mapping}
Map is an important prior knowledge the robot can leverage in an environment. From the perspective of an agentic framework, the map functions as an integral part of the robot's memory which allows the robot to efficiently determine object positions, accurately localize itself, and plan paths efficiently. Similar to \cite{mobilityvla}, we assume a demonstration tour video is given for the environment where the video can be taken from the robot or any other device. 

We propose an offline approach to construct a hybrid topological-semantic graph \( G = (\mathcal{V}, \mathcal{E}, \mathcal{L}) \) as our map representation that integrates geometric poses, visual landmarks, and connectivity constraints. Here, \(\mathcal{V}\) represents the set of nodes in the graph. \(\mathcal{E}\) is the set of undirected edges established between nodes based on their relative pose relationships, which is crucial for global path planning. The set \(\mathcal{L}\) holds the landmark information. For each visual landmark in the environment, \(\mathcal{L}\) records landmark details and all the node IDs where the landmark appears, effectively creating a centralized registry of its spatial occurrences. This allows the robot to utilize these landmarks for more accurate localization and navigation. 

The map construction process has three parts. \textbf{Topological map construction} sets up \(\mathcal{V}\) and defines \(\mathcal{E}\) based on relative poses for navigation. \textbf{Landmark semantic enrichment} extracts landmark details from nodes in \(\mathcal{V}\) to enhance semantic understanding. \textbf{Landmark co-visibility graph construction} perform cross-frame analysis to identify shared landmark features among multiple nodes, ensuring semantic consistency across the map.

\begin{figure*}[!ht]
\centering
\includegraphics[width=0.9\textwidth]{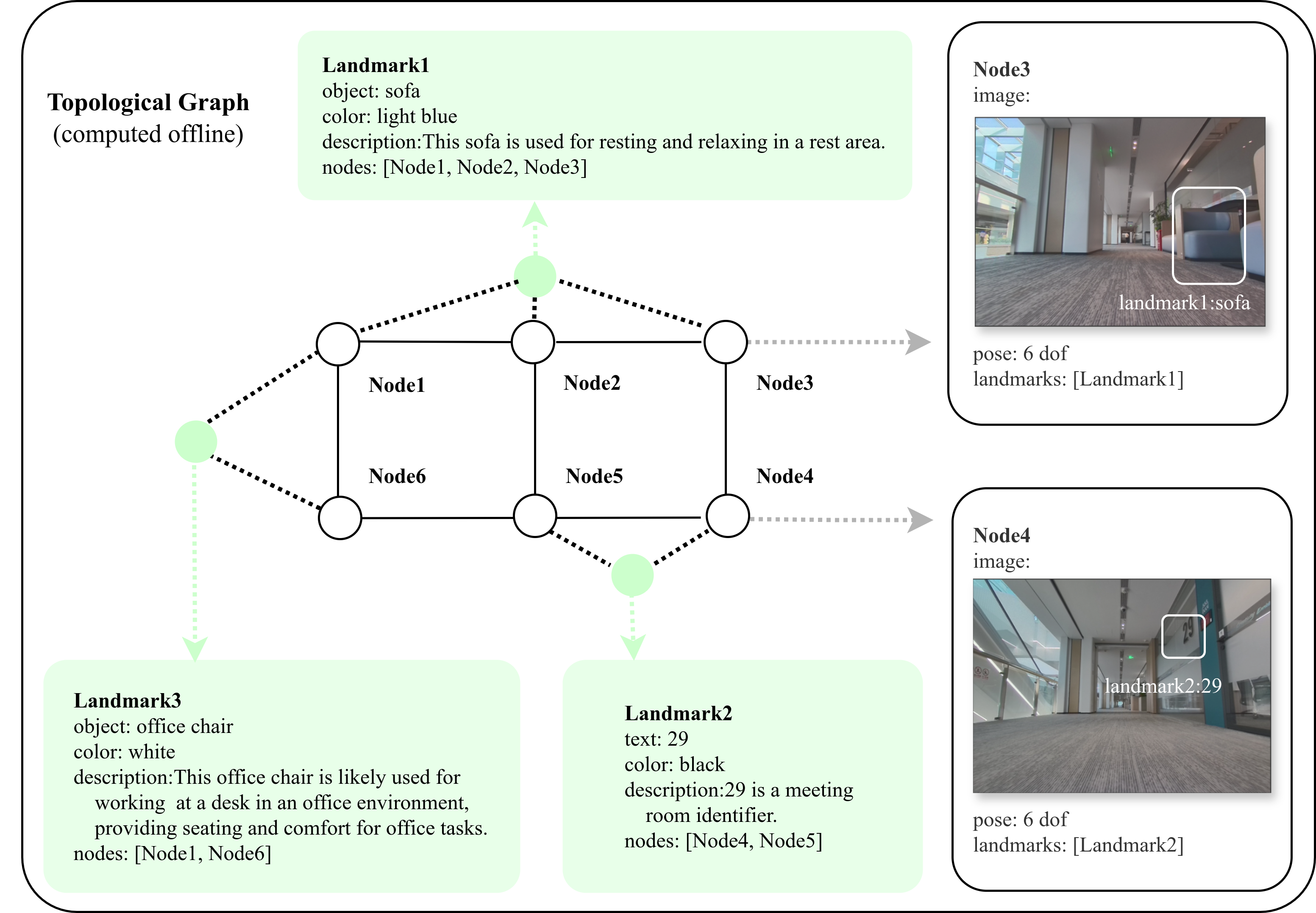}
\caption{Hybrid topological-semantic map structure. Nodes encode camera poses and landmark references; edges represent geometric connectivity; landmarks store semantic attributes and link to multiple nodes via co-visibility relationships.}
\label{fig:map_structure}
\end{figure*}

\textbf{Topological Map Construction:}  
The input video first undergoes temporal downsampling to reduce redundancy while preserving structural continuity. We employ \cite{schonberger2016structure}, an off-the-shelf structure-from-motion pipeline, to estimate approximate 6-Degree-of-Freedom (DoF) camera poses $\mathbf{T} \in SE(3)$ for each processed frame. The keyframes selected subsequently serve as nodes $\mathcal{V}$ in the hybrid map. Undirected edges $\mathcal{E}$ are established between nodes based on their relative pose relationships, thereby enabling global path planning and related navigation tasks. 

\textbf{Landmark Semantic Enrichment:}
\method-Global is employed to extract semantic landmarks from each node’s visual data, thereby enriching the map with high-level environmental semantics. For a given node \(v_i\), \method-Global combines linguistic guidance and image to identify a set of landmarks \(L_i = \{l_{i,1}, l_{i,2}, \dots, l_{i,k}\}\), where each landmark \(l_{i,m} \in \mathcal{L}\) is characterized by the following attributes:  
\begin{itemize}
\item Object/Text Category: Semantic labels such as "sofa," "A-001," or "door", providing fundamental entity identification;  
\item Visual Attributes: Color (e.g., "gray," "brown"), material (e.g., "fabric," "wood") or background (e.g. white wall, glass door), capturing perceptual characteristics;  
\item Functional Description: Natural-language annotations describing usage purposes (e.g., "for resting in living areas", "for document storage"), supporting language-based goal localization tasks. 
\end{itemize}
These landmark attributes are primarily represented in textual form. The first two attribute categories (object/text and visual attributes) are utilized across all localization tasks, while functional descriptions are specifically designed for language-based goal localization tasks.

\textbf{Landmark Co-Visibility Graph Construction:} 
To ensure semantic consistency across the map, we use \method-Global to perform cross-frame analysis to identify shared landmarks features visible from multiple nodes. When the same landmark (e.g., a gray sofa in a living room for resting) is detected in nodes \(v_i\) and \(v_j\), a bidirectional co-visibility relationship is established through two operations: 
\begin{itemize}
\item Landmark-to-Node Association: The landmark’s entry in \(\mathcal{L}\) is updated to include all node IDs where it appears (e.g., \(\text{nodes} = [\text{node1}, \text{node2}, \text{node3}]\)), creating a centralized registry of its spatial occurrences;  
\item Node-to-Landmark Reference: Each node \(v_i\) records the landmark's unique ID in its \(\text{landmarks}\) field, forming a many-to-many association that records which landmarks are visible from each node.  
\end{itemize}
This mechanism enables spatial relationship inference between non-adjacent nodes - for instance, determining navigable paths between rooms through shared landmarks like "door", even without direct geometric connections.

The resulting hybrid map integrates both geometric navigability (through topological edges) and semantic understanding (via landmark annotations), establishing a comprehensive prior knowledge for robotic navigation. As illustrated in Fig.~\ref{fig:map_structure}, this dual-representation architecture empowers the robot to: (1) localize itself or goal based on either language or image prompt, and (2) compute optimal navigation paths through the topological graph, thereby effectively bridging the semantic gap between high-level user instructions and low-level physical movement.

\subsubsection{Self \& Goal Localization}
Given the map defined above, \method-Global supports multi-modal localization requests, addressing two core tasks: vision-language localization and language-based goal localization as shown in Fig.~\ref{fig:astra_global}.

\textbf{Visual-Language Localization:}
\method-Global estimates an input image's pose in the map via a coarse-to-fine two-stage process. Coarse localization shrinks the pose search space, and then fine localization refines the result for higher-precision localization. 

In the coarse localization stage, the input to \method-Global consists of query image, localization prompt, and the pre-built landmark map. It analyzes the input image and takes into account the localization prompt to detect landmarks and establish correspondences with the pre-built landmark map. The model outputs detected landmarks in the query image : \( L_{\text{query}} = \{l_j | l_j = (\text{type}, \text{color} \ \text{or}\ \text{background})\} \) together with the matched landmarks \( \mathcal{L}_{\text{matched}} \subseteq \mathcal{L} \). In this step, the model mainly focuses on semantic matching that considers both categorical alignment (e.g., "sofa" $\leftrightarrow$ "couch") and attribute consistency (color/texture/background similarity). For example, a query for a "gray sofa" retrieves map nodes containing corresponding "light-gray couch" entries via the \method-Global's semantic reasoning. 

However, the landmark results obtained above are merely matching results based on the text descriptions of landmarks. To further optimize the coarse localization process, it is crucial to also consider image similarity. We term this step the visual consistency filtering. Specifically \method-Global determines whether there is a co-visible area between a candidate image corresponding to each landmark and the query image. Landmarks without a co-visible area are filtered out. This additional step refines the set of matched landmarks \( \mathcal{L}_{\text{filtered}}\), ensuring that the results are more accurate and relevant to the actual visual content of the query image, thus enhancing the effectiveness of the coarse localization process.

The final candidate nodes are determined by:
\[
\mathcal{V}_{\text{candidate}} = \bigcup_{l_k \in \mathcal{L}_{\text{filtered}}} \{v_i | v_i \leftrightarrow l_k \}.
\]




In the fine localization stage, the goal is to leverage the query image and \(V_{candidate}\) output from the coarse localization to get a more precise localization result. To bridge the gap between coarse candidate regions and accurate pose estimation, we sample reference map nodes \(V_{ref}\) from offline map \(G\) that are in close proximity with \(V_{candidate}\). Proximity is simply measured by the Euclidean distance between the location and the difference between the pose. \(V_{ref}\) serves as anchor points that allows for more detailed comparison with the query for better localization accuracy.





With \(V_{ref}\) established, we feed both the query image and reference nodes (each containing an image and its known pose) to \method-Global which directly outputs the predicted pose of the query image by leveraging the combined visual and positional information from the reference nodes. The prediction is formulated as: 
\begin{equation}
\hat{p} = \text{Astra-Global}\left(I_{query}, V_{ref} \right) .
\end{equation}

\textbf{Language-based Goal Localization:}
In this task, the overarching goal is to retrieve the image of the target location within the map \(\mathcal{G}=(\mathcal{V},\mathcal{E},\mathcal{L})\) that corresponds to the given natural language instruction. As landmarks in our map \textit{G} already contains functional description, it's straightforward to use \method-Global to identify the relevant landmarks $l \in \mathcal{L}$ that satisfy the query language instruction. Subsequently, through the landmark-to-node association mechanism, we can locate the relevant nodes. The nodes, in turn, provide the images and 6-DoF poses $\mathbf{T}_v \in SE(3)$  that contain the desired goal.

To ensure efficient retrieval, the system implements a spatial partitioning strategy that divides the search space \(\mathcal{S}\) into sub-regions \(\{\mathcal{S}_1,\mathcal{S}_2,\ldots,\mathcal{S}_n\}\). Formally, using a Euclidean distance metric \(d(\cdot,\cdot)\) in 3D space, the retrieval process first focuses on landmarks \(l\) within sub - region \(\mathcal{S}_j\) satisfying \(\forall \mathbf{p} \in \mathcal{S}_j, d(\mathbf{p},\mathbf{p}_r)\leq r\), where \(r\) represents a predefined search radius. Initially, the system gives priority to searching the area near the current location. If the target location image is not found in this nearby area, the search will then be expanded to the surrounding regions. This process enables the system to narrow down the potential locations within the map, with the ultimate goal of accurately identifying the image of the target location as specified by the user's natural language input.

\subsubsection{Model Training}
\method-Global leverages Qwen2.5-VL \cite{bai2025qwen2} as its backbone, combining supervised fine-tuning (SFT) and Group Relative Policy Optimization (GRPO)\cite{shao2024deepseekmath} to specialize the model for localization tasks while retaining its general multimodal capabilities.

In the SFT stage, we prepare diverse datasets with different tasks to fine-tune the model. Besides coarse and fine localization datasets as the main tasks, we also constructed a series of auxiliary tasks to help improve model's spatial understanding including:
\begin{itemize}
    \item Co-Visibility Detection: given two images, the model must determine whether a target image and a reference image share a co-visible region.
    \item Selection of Co-Visible Image: the model needs to identify the image from several candidate images that has co-visibility with the query image.
    \item Estimation of Movement Trend: given two images, the model needs to infer the relative movement trend from a reference image to a target image.
\end{itemize}

These auxilary tasks fosters the model’s general understanding of spatial relationships and task constraints, enabling it to derive actionable decisions through sequential reasoning rather than direct input-output mapping. Notably, all datasets are engineered to minimize annotation overhead: data collection relies solely on image-pose pairs, a standard output of robotic SLAM systems or open-source datasets.

Following the SFT stage, we employ GRPO to train the model for visual-language localization tasks, leveraging rule-based reward functions tailored to coarse localization subtasks. 
To facilitate the training of reinforcement fine-tuning, the model formats the obtained results from the coarse localization process into a predefined, structured output. This format is specifically designed to simplify the calculation of the reward. 
The reward function $R_{course}$ consists of four normalized components.
\[  
R_{\text{coarse}} =  R_{\text{format}} +  R_{\text{landmark}} +  R_{\text{map}} + R_{\text{extra}},  
\]
where each reward is defined as follows:

\textit{Format Reward (\( R_{\text{format}} \))}: Checks if the output adheres to the predefined format:  
   \[  
   R_{\text{format}} = \begin{cases}  
   1 & \text{if format is valid}, \\  
   0 & \text{otherwise}.  
   \end{cases}  
   \] 
   
\textit{Landmark Extraction Reward (\( R_{\text{landmark}} \))}: Rewards accurate extraction of semantic landmarks by matching against ground-truth landmarks \( \mathcal{G}_{\text{landmark}} \):  
   \[  
   R_{\text{landmark}} = \frac{|\mathcal{P}_{\text{landmark}} \cap \mathcal{G}_{\text{landmark}}|}{|\mathcal{G}_{\text{landmark}}|},  
   \]  
   where \( \mathcal{P}_{\text{landmark}} \) is the set of predicted landmarks.
   
\textit{Map Matching Reward (\( R_{\text{map}} \))}: Computed as the Intersection-over-Union (IoU) between predicted and ground-truth landmark ids: 
   \[  
   R_{\text{map}} = \frac{|\text{ids}_{pred} \cap \text{ids}_{gt}|}{|\text{ids}_{pred} \cup \text{ids}_{gt}|}.  
   \]  
   
\textit{Extra Landmark Reward (\( R_{\text{extra}} \))}: Rewards novel landmarks that are not in the ground truth but correct, weighted by their pose error to query image's ground truth pose.
    \[  
   R_{\text{extra}} = \exp\left(-\lambda \left( w_d \cdot d(p_{\text{pred}}, p_{\text{gt}}) + w_\theta \cdot |\phi_{\text{pred}} - \phi_{\text{gt}}| \right)\right),
   \] 
   where \(d(.,.)\) is the Euclidean distance, \( w_d+w_\theta=1 \) are used to balance position and angular errors. 

While landmark-based coarse retrieval effectively narrows down candidate locations using semantic landmarks (e.g., objects and visual attributes), this process may inadvertently discard fine-grained visual features that are crucial for precise localization. To mitigate this limitation, we introduce a visual consistency filtering mechanism that evaluates co-visibility relationships between query images and retrieved landmark candidates. This step ensures that only candidates exhibiting both semantic relevance and visual consistency are advanced to the fine localization stage.

For the visual consistency filtering method, the reward combines two components: format consistency and co-visibility score consistency. The reward of co-visibility score consistency is computed as:

\begin{equation}
    R_{\text{covis}} = 1 - |S_{\text{gt}} - S_{\text{pred}}|,
\end{equation}

where:
\begin{itemize}
    \item $S_{\text{gt}}$ is the ground truth co-visibility score
    \item $S_{\text{pred}}$ is the predicted co-visibility score from model inference
\end{itemize}

The total reward for visual consistency filtering is then calculated as:

\begin{equation}
    R_{\text{total}} = R_{\text{format}} + \lambda R_{\text{covis}},
\end{equation}

where $\lambda$ is a weighting hyperparameter balancing the importance between format consistency and co-visibility consistency.

\subsection{\method-Local}
\label{sec:astra_local}

\begin{figure*}[h!t]
\centering
 \includegraphics[width=1\textwidth]{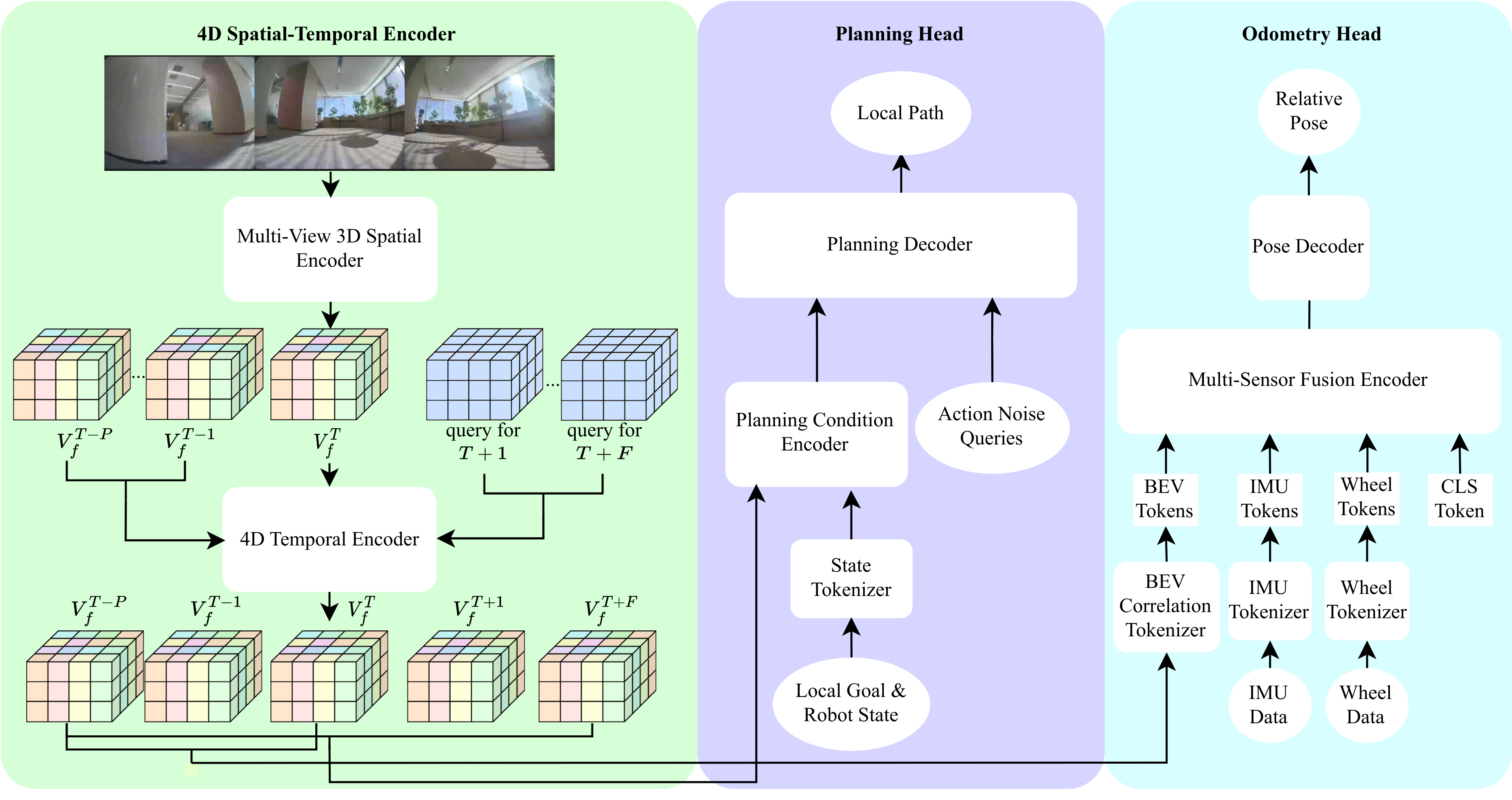}
\caption{Model Architecture of \method-Local. The multi-view images first go through a 3D spatial encoder to get a voxel feature $V_f^T$ for current frame. Combined with voxel features from previous frames and query embeddings for future frames, 4D Temporal Encoder predicts voxel features for future timestamps. The Odometry head incorporates the current and prior voxel features along with additional sensor data. it leverages a transformer encode to fuse multiple modalities and outputs a relative pose. The planning head takes all predicted voxel features with local goal and other robot states as inputs and formulates local path planning as conditional flow matching.}
\label{fig:astra_local}
\end{figure*}

\method-Local is a multi-task network that generates local paths and estimates odometry from sensor data in an end-to-end fashion. As shown in Fig.~\ref{fig:astra_local}, the architecture of \method-Local comprises three main components. First, the 4D Spatial-Temporal encoder processes multi-frame, multi-camera inputs to generate 4D features aligned with real-world coordinates, which serve as the foundation for downstream tasks. Second, the Planning Head takes these 4D features along with local goal and robot status and generates executable trajectories for the robot to follow. Third, the Odometry Head uses the same 4D features with additional sensor inputs to estimate the relative pose between the current frame and previous frames.

\subsubsection{4D Spatial-Temporal Encoder}
In traditional mobility stack, perception and prediction modules serve as critical components by enabling agents to fuse information from the past and acquire both current and future environmental states. In \method-Local, we propose a unified 4D spatial-temporal encoder to supplant the two distinct modules. The 3D spatial encoder is trained at first to generate common representations through vast amounts of unlabeled data via a self-supervised learning paradigm. Then, the 4D spatial-temporal encoder is trained to forecast future environmental representations on top of the 3D encoder.

\textbf{3D Spatial Encoder:}  
Given $N$ surround-view images, the 3D encoder $\mathcal{M}_{3D}$ is designed to encode them into geometric 3D representations:
\begin{equation}
    \mathbf{V_f} = \mathcal{M}_{3D}(\sum_i^N{  \mathbf{I}_i, \mathbf{K}_i, \mathbf{T}_i) },
\end{equation}
where $\mathbf{I}_i$, $\mathbf{K}_i$, and $\mathbf{T}_i$ correspond to the $i$th image and its intrinsic and extrinsic. $\mathbf{V}$ denotes the voxel-level 3D representations.

Specifically, we employ a Vision Transformer (ViT) \cite{dosovitskiy2020vit} to encode input images into discriminative feature representations $\mathbf{F}_i \in \mathcal{R}^{h \times w \times C}$ from the $i$th image $\mathbf{I}_i \in \mathcal{R}^{H \times W \times 3}$. Then, we use lift-splat-shoot \cite{philion2020lift} to convert 2D image features into 3D voxel features:
\begin{equation}
    \mathbf{V}_{f} = \mathcal{P}  (\sum_i^N{  \pi (\mathbf{F}_i \otimes \mathbf{D}^{dist}_{i}, \mathbf{K}_i, \mathbf{T}_i)) },
\end{equation}
where $\mathbf{D}^{dist}_{i}$ represents the estimated depth distribution. $\pi$ refers to the projection which transforms a 2D image pixel into a 3D point using the camera intrinsic $\mathbf{K}$ and extrinsic $\mathbf{T}$. The function $\mathcal{P}$ denotes voxel pooling.

We train the 3D Spatial Encoder in a self-supervised learning manner and is achieved via 3D volumetric differentiable neural rendering where the depth and color image are rendered from $\mathbf{V_f}$. Specifically, to render the depth image, we use a multi-layer perceptron (MLP) \cite{haykin1994neural} to transform the voxel feature $\mathbf{V}_f$ into a signed distance field (SDF) $\mathbf{V}_s$. Then, given a set of rays $\mathbf{r}$ consisted of camera origins and view directions, the depth values are computed through a weighted integration along the rays. The weighting coefficients are derived from the opacity and the accumulated transmittance as in MonoSDF. Similarly, we apply an MLP to $\mathbf{V_f}$ to get $\mathbf{V_{color}}$ and use the same logic to render a color image from $\mathbf{V_{color}}$. For the $i$th image, the neural rendering is represented by:
\begin{equation}
\mathbf{D}^{\mathbf{r}}_i = \mathcal{R}(\mathbf{V}_s, \mathbf{r}_i), \quad\mathbf{I}^{\mathbf{r}}_i = \mathcal{R}(\mathbf{V}_{color}, \mathbf{r}_i).
\end{equation}

The ground truth depth labels are needed for supervision along with the original color images. For certain open-source datasets, depth are provided. However, in real-world scenarios, depth labels are typically unavailable. To address this limitation, we leverage large-scale mono depth estimation model, \ie DepthAnything-V2 \cite{yang2024depth}, to generate pseudo depth labels. The pseudo depth labels are subsequently aligned with depth measurements from depth sensors such as Lidar or RGBD cameras if available through RANSAC, to produce dense and accurate depth maps for training. The loss function in a batch can be formulated as: 
\begin{equation}
    \mathcal{L}_{3D} = \frac{1}{N} \sum_i^N (|| \mathbf{D}^{gt}_i(\mathbf{r}_i) -   \mathbf{D}^{\mathbf{r}}_i||_1 + || \mathbf{I}_i(\mathbf{r}_i) -   \mathbf{I}^{\mathbf{r}}_i||_1).
    \label{eq:3dpretrainloss}
\end{equation}
$\mathbf{D}^{gt}_i(\mathbf{r}_i)$ and $\mathbf{I}_i(\mathbf{r}_i)$ represent the ground truth depth values and color values sampled by $\mathbf{r}_i$.

\textbf{4D Spatial-Temporal Encoder:}
$\mathbf{V}_f$ representing the current environment serves perception clues for planning where $\mathbf{V}_s$ can be easily converted into occupancy $\mathbf{V}_o$ based on the sign of an SDF value. However, besides perception, prediction is also critical for local planning and it models the temporal dynamics of environmental evolution. To this end, we propose a model that directly forecasts future voxel-level spatiotemporal representations. 

Specifically, the 4D encoder $\mathcal{M}_{4D}$ takes the past voxel features and future timestamps as input, and outputs future voxel features at corresponding timestamps: 
\begin{align}
    &\{ \hat{V}_f^j | j=T+1, \cdots, T + F \} = \nonumber \\  
    &\mathcal{M}_{4D}(\{ V_f^j | j=T-P, \cdots, T \}, \{ t_j | j=T+1, \cdots, T + F\}).
\end{align}
where $t_j$ denotes the timestamp of the $j$th frame.
The past and current $P + 1$ voxel features are concatenated in channels and encoded into multi-scale features by a ResNet, which are composed of 3D convolutions. The prediction module is inserted at each feature level. It is implemented by DiT \cite{peebles2023scalable} blocks, where encoded features serve as input while future timestamps serve as condition. Then, we adopt an FPN3D to fuse the multi-scale predicted features and obtain final future $F$ voxel features.

The prediction module is also trained in a self-supervised fashion following the pretraining of the 3D Spatial Encoder. The loss function is computed at both voxel level and pixel level:
\begin{align}
\mathcal{L}_{4D} = & \frac{1}{F} \sum_j^F ( ||V_f^j - \hat{V}_f^j||_1)  + \nonumber \\
&\frac{1}{FN}\sum_j^F\sum_i^N( || \mathbf{D}^{gt}_{ij}( \mathbf{r}_{ij}) - \mathcal{R}(\hat{V}_s^j, \mathbf{r}_{ij}) ||_1  + \nonumber \\
&|| \mathbf{I}_{ij}( \mathbf{r}_{ij}) - \mathcal{R}(\hat{V}_{color}^j, \mathbf{r}_{ij}) ||_1 ).
\label{eq:4dloss}
\end{align}
The first term measures the voxel-level differences between the original and predicted future voxel features. The second term is computed at the pixel level, where $\mathbf{D}^{gt}_{ij}( \mathbf{r}_{ij})$ and $\mathbf{I}_{ij}( \mathbf{r}_{ij})$ represent the ground truth depth values and color values sampled from the $i$ th view of the $j$ th frame by a set of rays $\mathbf{r}_{ij}$, respectively. $\hat{V}_s^j$ and $\hat{V}_{color}^j$ are obtained from the predicted voxel features $\hat{V}_f^j$ via MLPs. 

After pretraining, the full 4D Sptial-Temporal Encoder is able to produce both current and future environmental states, represented by 4D voxel features. The framework is trained in a self-supervised manner where only depth labels are needed. Also note that the model design and training procedure can support different number of input views which means besides our own data, we can also leverage diverse open-source datasets like depth estimation, autonomous driving and etc.

\subsubsection{Planning Head}
For our local path planning problem, we define the action trajectory in the format of 
\begin{equation}
    X = <(\Delta x_1, \Delta y_1, \Delta \theta_1), \cdots, (\Delta x_n, \Delta y_n, \Delta \theta_n)> ,
\end{equation}
where each $(\Delta x_i, \Delta y_i, \Delta \theta_i)$ represents the relative coordinates between adjacent poses.
Our planning head takes the pretrained 4D features as condition, along with robot velocity and task information like goal pose, and reconstructs an action trajectory from a Gaussian noise with flow matching \cite{LipmanY23}. 

\textbf{Transformer-based flow matching}:
Given that our planning task deals with complex environments where trajectories often have multi-modal characteristics, generative methods like diffusion models and flow matching are better suited as are also commonly used in autonomous driving and robot manipulation \cite{chi2023diffusion, black2024pi_0, liao2024diffusiondrive}.

We choose flow matching as our main planning method due to its high efficiency which is vital to a real-time system.
More specifically, we aim to minimize the Conditional Flow Matching (CFM) objective \cite{LipmanY23}: 
\begin{equation}
    \mathcal{L}_{CFM}(\theta) = \mathbb{E}_{t, X_1, X_t}\|v_t(X_t; \theta) - u_t(X_t|X_1)\|^2,
\end{equation}
where $\theta$ is the model parameter, $t \in U[0, 1]$ is the timestep, $X_0 \in p$ is the source distribution (usually a normal distribution), $X_1 \in q(\cdot|C)$ is the target distribution and $C$ is the condition mentioned above.

A transformer-based model is used to represent the vector field, \ie $v_t(X; \theta |t, C)$. 
The condition $C$ consists of four parts: robot velocities, goal points, voxel features $\{V_f^j\}$ and occupancy maps $\{V_o^j\}$. 
Robot velocities and goal points are normalized and projected to the robot's ego coordinate respectively, and then tokenized by a linear projection. 
We concatenate the channel and time dimensions of voxel features, and further combine it at channel dimension with occupancy maps encoded by a convolution neural network. 
The combined features are tokenized by unfolding its spatial dimensions (width and height) and a 2D positional embedding is also applied.
Afterwards, a transformer encoder takes all the tokens above, as well as the timestep features as inputs, and sends the results to a transformer decoder, followed by an MLP-based action head that outputs the corresponding vector.

\textbf{Masked ESDF loss}:
A major challenge of our task is to avoid collision with various types of obstacles (static or dynamic) in the environment.  
Although some guidance-based techniques have been developed \cite{FengR25} to control the vector field of flow matching with a cost function, they may also introduce additional computational overhead, \eg either conducting Monte Carlo estimation at inference time \cite{ChenL24}, or learning a surrogate model at training time \cite{LuC23}.

To achieve the required low inference latency in our real-time system while maintaining the training cost unchanged, we design a novel technique named masked ESDF loss.
Specifically, given a 3D occupancy map $V_o$, we calculate its euclidean space distance field (ESDF) map $\Phi(x, y)$ by: \textit{1).} compressing the 3D map to a 2D binary map $V_{m}(x, y)$ by taking maximum on the z-dimension, \textit{2).} calculating the minimum Euclidean distance from each free pixel to the nearest obstacle pixel \cite{FelzenszwalbH12}:
\begin{equation}
\Phi(x,y) = 
\begin{cases} 
+D(x,y) & \text{if } V_{m}(x,y) = 0 \text{ (free space)} \\
-D'(x,y) & \text{if } V_{m}(x,y) = 1 \text{ (obstacle)}
\end{cases}
\label{eq:esdf}
\end{equation}
where $D$ and $D'$ represent the unsigned Euclidean distance transform and the interior distance transform respectively.

During the training phase of flow matching, we approximate the action trajectory with the predicted vector $v_t$ by:
\begin{equation}
    \tilde{X} = X_t - t \cdot v_t(X_t; \theta | t, C).
\end{equation}
We further calculate the pose trajectory $\tilde{X}_p$ based on:

\begin{align*}
    \left\{
        \begin{aligned}
        \begin{bmatrix}
        x_k \\ y_k 
        \end{bmatrix}
        & = \begin{bmatrix}
        x_{k-1} \\ y_{k-1} 
        \end{bmatrix}
        +
        \begin{bmatrix}
        \cos\theta_{k-1} & -\sin\theta_{k-1} \\
        \sin\theta_{k-1} & \cos\theta_{k-1} \\
        \end{bmatrix}
        \begin{bmatrix}
        \Delta x_k \\ \Delta y_k \\
        \end{bmatrix} \\
        \theta_k &= \theta_{k-1} + \Delta\theta_k
        \end{aligned}
    \right.
    \label{eq:system}
\end{align*}

Then the ESDF value of each point on the trajectory can be queried by $\Phi(x, y), (x,y) \in \tilde{X}_p$.

However, directly minimizing the ESDF values will result in heading errors, \ie the trajectory will always head to areas with fewer obstacles rather than the areas of goal points.
To address this issue, we add a 2D ground-truth trajectory mask on the ESDF map:
\begin{equation}
    \tilde{\Phi}(x,y) = \Phi(x, y) \cdot (1 - \alpha \cdot \mathbb{I}((x,y)\in U_{gt})),
\end{equation}
where $U_{gt}$ is the expanded area of the ground-truth trajectory and we use $\alpha$ to punish reconstructed trajectories that are too far away from the ground-truth. 

The total loss function with masked ESDF loss can be written as:
\begin{equation}
    \mathcal{L}_{planning} = \mathcal{L}_{CFM} - \lambda \cdot \sum_{(x,y) \in \tilde{X}_p} \tilde{\Phi}(x,y).
\end{equation}
In our implementation, the masked ESDF loss is calculated by bilinear grid sampling.

The proposed masked ESDF loss only introduces $O(n)$ additional computational overhead where $n$ is the size of action trajectory, and brings no extra cost to the inference time. We demonstrate in experiments that the collision rate can be significantly reduced with this proposed loss. 

\subsubsection{Odometry Head}

The odometry head predicts relative robot pose given current and past 4D features from the 4D encoder and additional sensor including IMU, wheel. We train a transformer model to fuse information from different sensors (see Fig.~\ref{fig:astra_local}). For each time step, each modality (4D vision feature, IMU, wheel) goes through a modality-specific tokenizer to get tokens for that time step. Combined with learned modality embedding and temporal position embedding, the tokens from a series are fed into a transformer encoder together with a CLS token which is then used to predict the relative robot pose. Specifically, for 4D vision feature tokenizer, we took inspiration from \cite{bevodom} and computed a correlation volume between two consecutive 3D voxel features. For IMU and wheel data, we use a small LSTM to encode raw data between two image frames into a single token. Training for the odometry head follows a straightforward supervised learning setup where the objective is to minimize the L1 loss between the predicted delta pose and ground truth pose. 
\section{Experiments}
\label{exp}

We collected data from our in-house built robots operating in diverse environments such as warehouses, office buildings, and homes to train and test \method. We optimized and deployed \method on our robots. Specifically, \method-Local runs on the on-bot edge device, while \method-Global runs on the cloud.

\begin{numberedconclusion}
\method achieves high end-to-end mission success rate across diverse environments. 
\end{numberedconclusion}

To assess the end-to- end performance of \method, we adopt the settings from \cite{mobilityvla}, where user instructions are randomly selected and the robot is initially placed at random locations within the environment, and then measure the overall success rate of the robot reaching the goal. Additionally, we evaluate the performance of \method for three navigation sub-tasks, \ie goal localization, self-localization, and path planning.
For goal localization, similar to \cite{mobilityvla}, we measure the success rate (SR) of \method in correctly localizing user instructions.
For self-localization, we compare the estimated robot trajectory with the ground-truth trajectory and consider it a success if the translational error for the entire trajectory is within 1m.
For path planning, we implemented a fallback system on the robot. When the trajectory generated by \method-Local violates collision constraints, the system falls back to a traditional optimization-based planning method. We report the fallback rate (FR) as an indicator of how often the system relies on this fallback mechanism.

\begin{table}[!h]
    \centering
    \begin{tabular}{|c|c|c|} \hline 
         &  Warehouse&  Office Building\\ \hline 
 End-to-end SR& 84.2\%  & 99.1\% \\\hline
         Goal Localization SR& 98.3\%  & 99.1\%  \\ \hline 
         Self-Localization SR&  85.9\%& 100\%  \\ \hline 
         Path Planning FR& 8.3\% & 15.6\% \\ \hline
    \end{tabular}
    \caption{Success Rate (SR) of \method for end-to-end mission, goal localization, self-localization and Fallback Rate (FR) for path planning across diverse environments}
    \label{tab:astra_sr}
\end{table}

As shown in Tab.~\ref{tab:astra_sr} \method achieves a high end-to-end success rate in all types of environments. Among the failures, the primary reason for failure in warehouses is due to the fact that at the starting point the robot is not able to self localize. This is because of the highly repetitive environment and the lack of visible landmarks around some random starting locations. When excluding these challenging starting points, the success rate increases to 91.2\%. In the office building environment, although the end-to-end success rate is high, the path planning fallback rate increases due to the presence of more dynamic obstacles.

\begin{numberedconclusion}
\method-Global can handle multi-modal localization query across diverse environment. 
\end{numberedconclusion}

\begin{figure}[!htpb]
\centering
\includegraphics[width=1\textwidth]{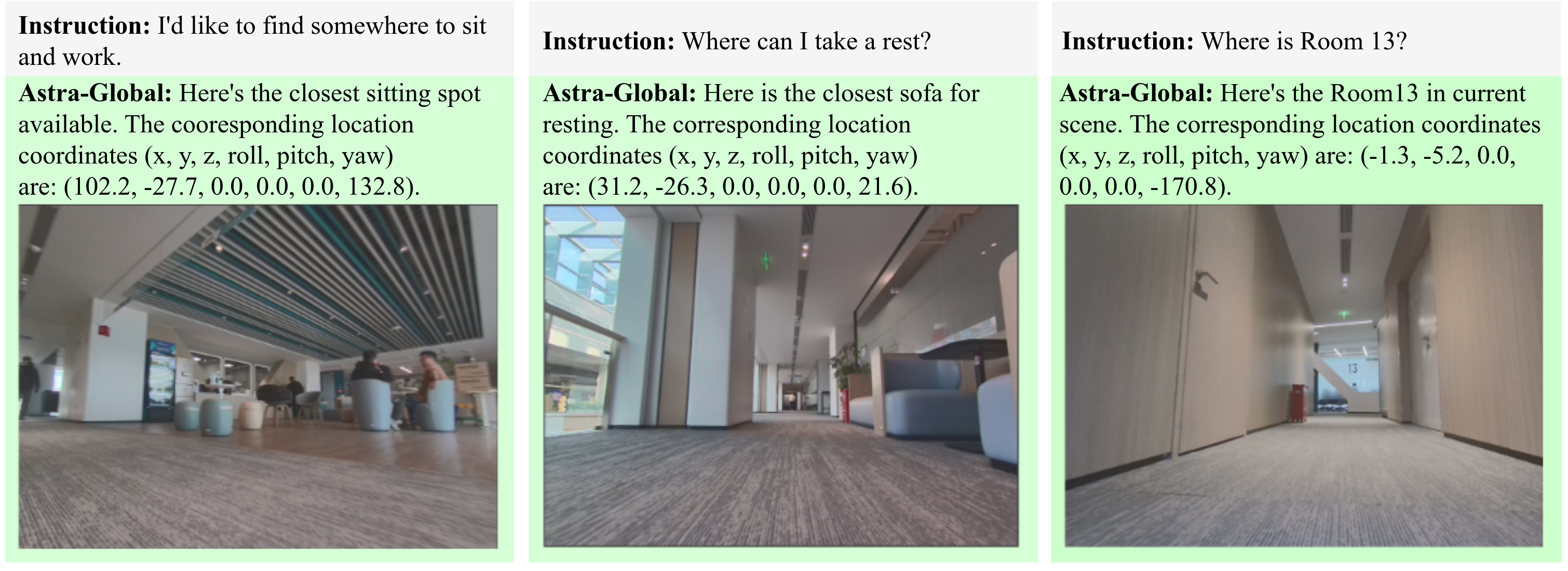}
\caption{Language-based goal localization task examples.}
\label{fig:instruction_example}
\end{figure}

\begin{figure}[!htbp]
    \centering
    \includegraphics[width=1\textwidth]{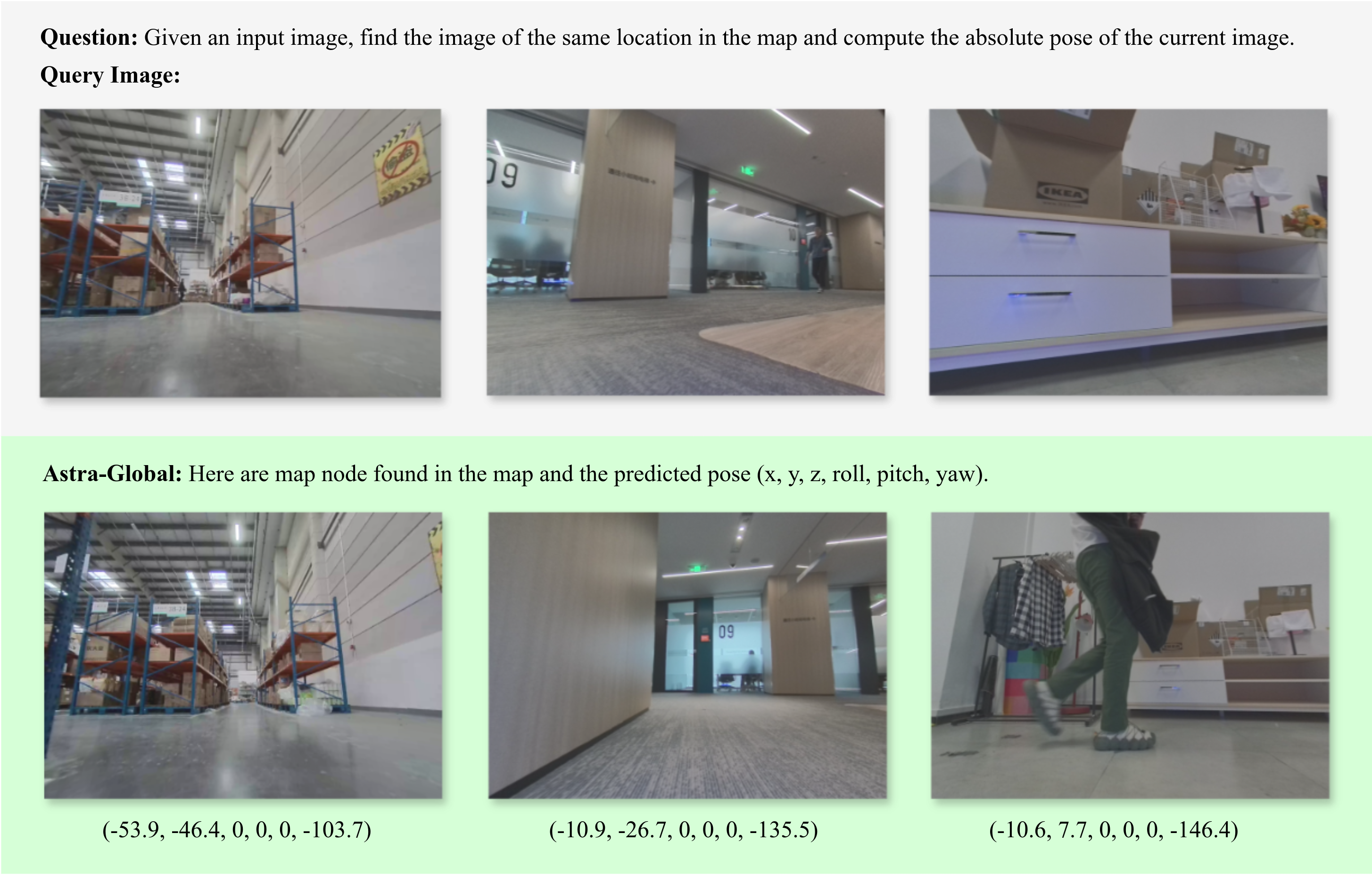} 
    
    \vspace{6pt} 
    \makebox[\textwidth]{%
        \makebox[0.27\textwidth]{\hfill(a) Warehouse Localization\hfill}%
        \makebox[0.27\textwidth]{\hfill(b) Office Localization\hfill}%
        \makebox[0.27\textwidth]{\hfill(c) Home Localization\hfill}%
    }
    \caption{Self-Localization across diverse scenarios.}
    \label{fig:reloc_example_all}
\end{figure}

With our map representation, \method-Global supports text and image localization queries, as illustrated in Fig.~\ref{fig:instruction_example},~\ref{fig:reloc_example_all}. For goal localization, \method-Global can effectively identify map images and poses matching text instructions.

\begin{figure}[!htbp]
    \centering
    \includegraphics[width=0.6\textwidth]{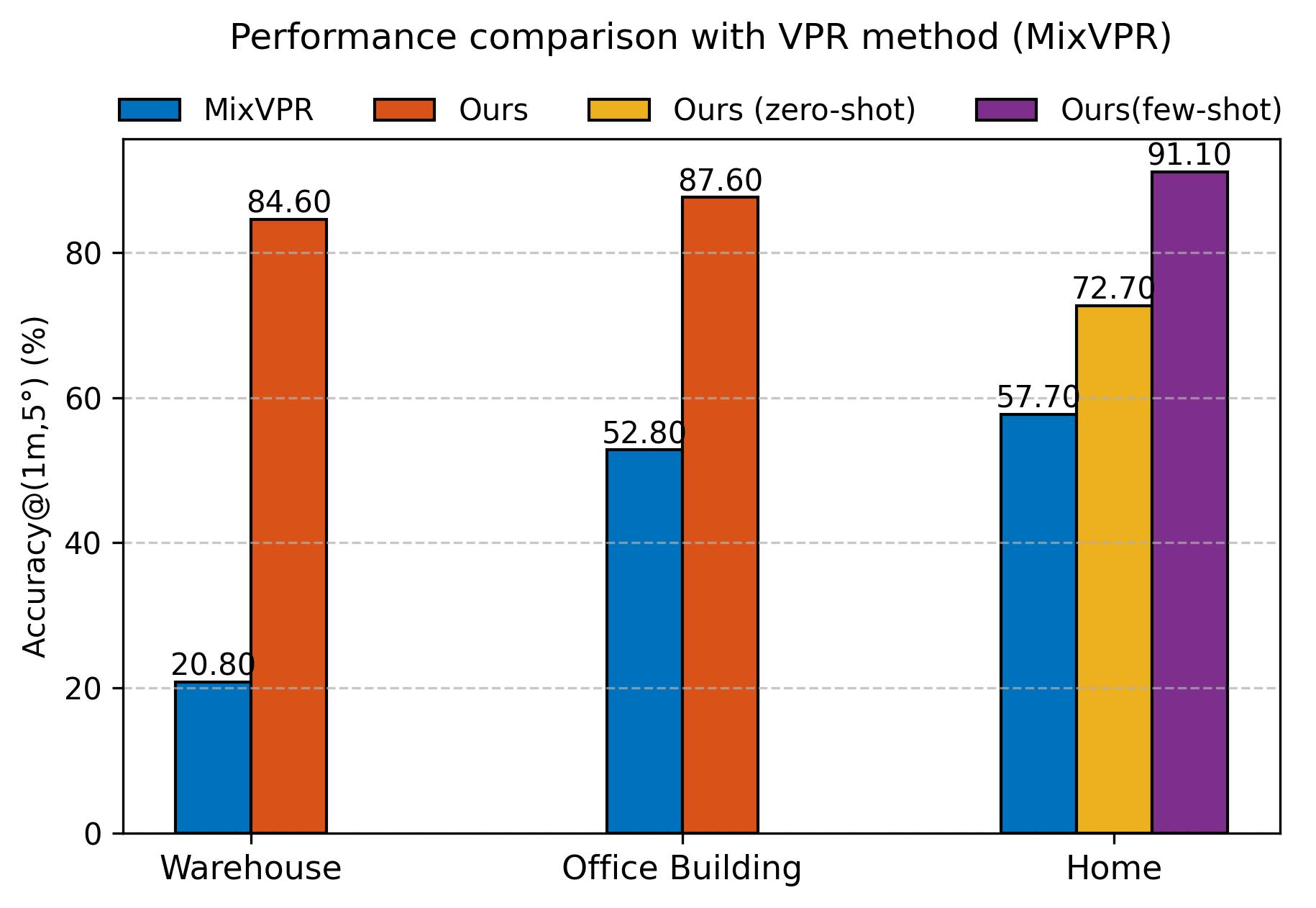} 
    \caption{Performance (\%) comparison with VPR method (MixVPR) across diverse indoor environments. }
    \label{fig:compare_with_vpr}
\end{figure}

For robot localization, \method-Global showcases excellent performance across a wide range of scenarios. As Fig.~\ref{fig:compare_with_vpr} and Fig.~\ref{fig:reloc_example_all}(a)(b) show, whether tested in large-scale industrial-looking warehouses or visually distinct office buildings, the model demonstrates its effectiveness. It can adapt well to the significant scale and visual differences between these diverse scenarios, delivering reliable localization results.

We compare \method-Global with traditional visual place recognition (VPR) method that often relies on models to provide an image embedding and formulate the problem as retrieval. To make a fair comparison between our method and traditional VPR methods, we ensure that the recall rates of the two methods are the same and focus on comparing their precision. The evaluation metrics include the accuracy of pose within a 1-meter distance error and 5-degree angular error and the results are presented in Fig.~\ref{fig:compare_with_vpr}.

\begin{figure}[!h]
    \centering
    \begin{subfigure}[b]{0.7\textwidth}
        \includegraphics[width=\textwidth]{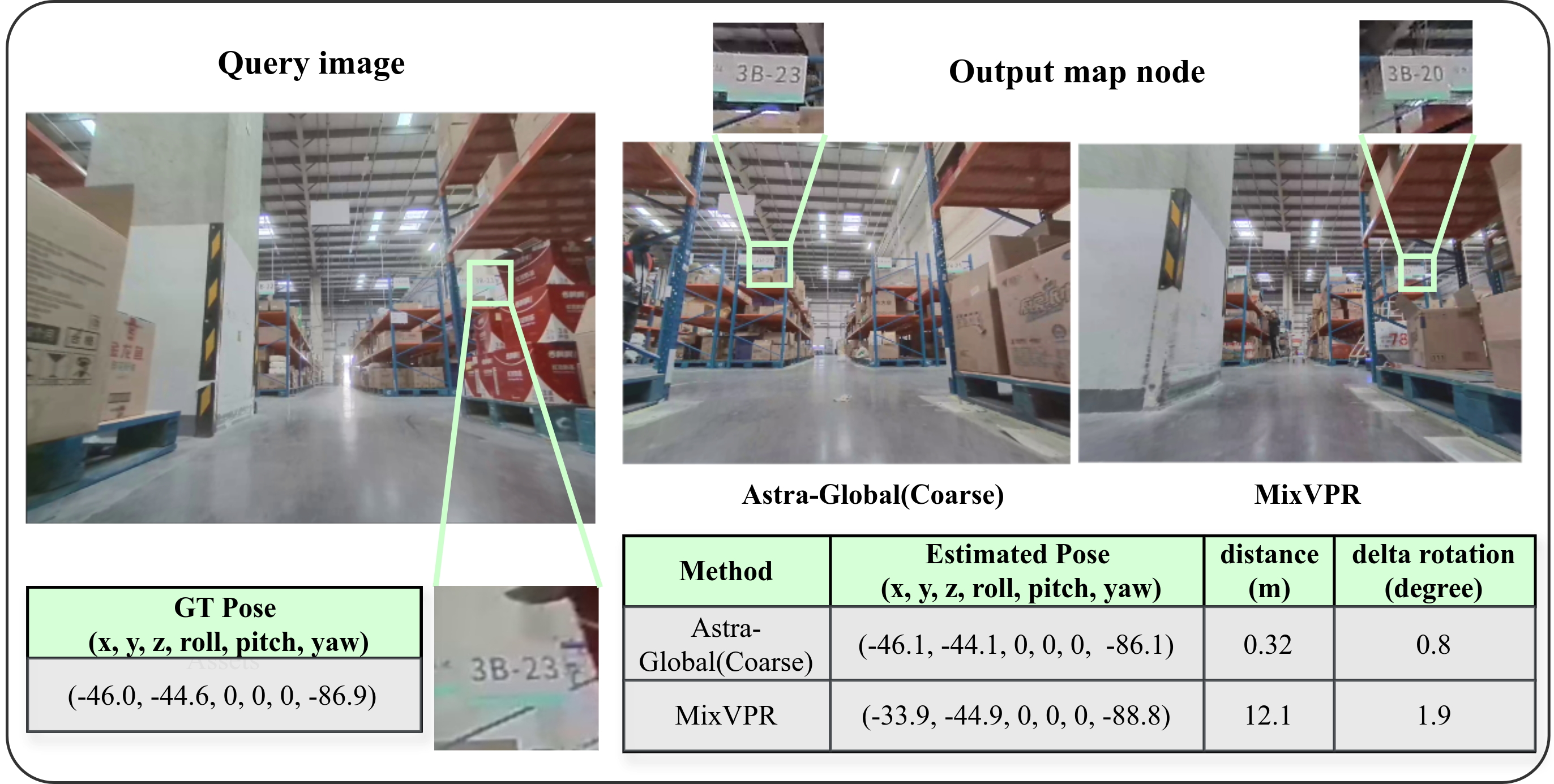} 
        \caption{Warehouse Localization} 
        \label{fig:compare_vpr_a} 
    \end{subfigure}
    \quad
    \begin{subfigure}[b]{0.7\textwidth}
        \includegraphics[width=\textwidth]{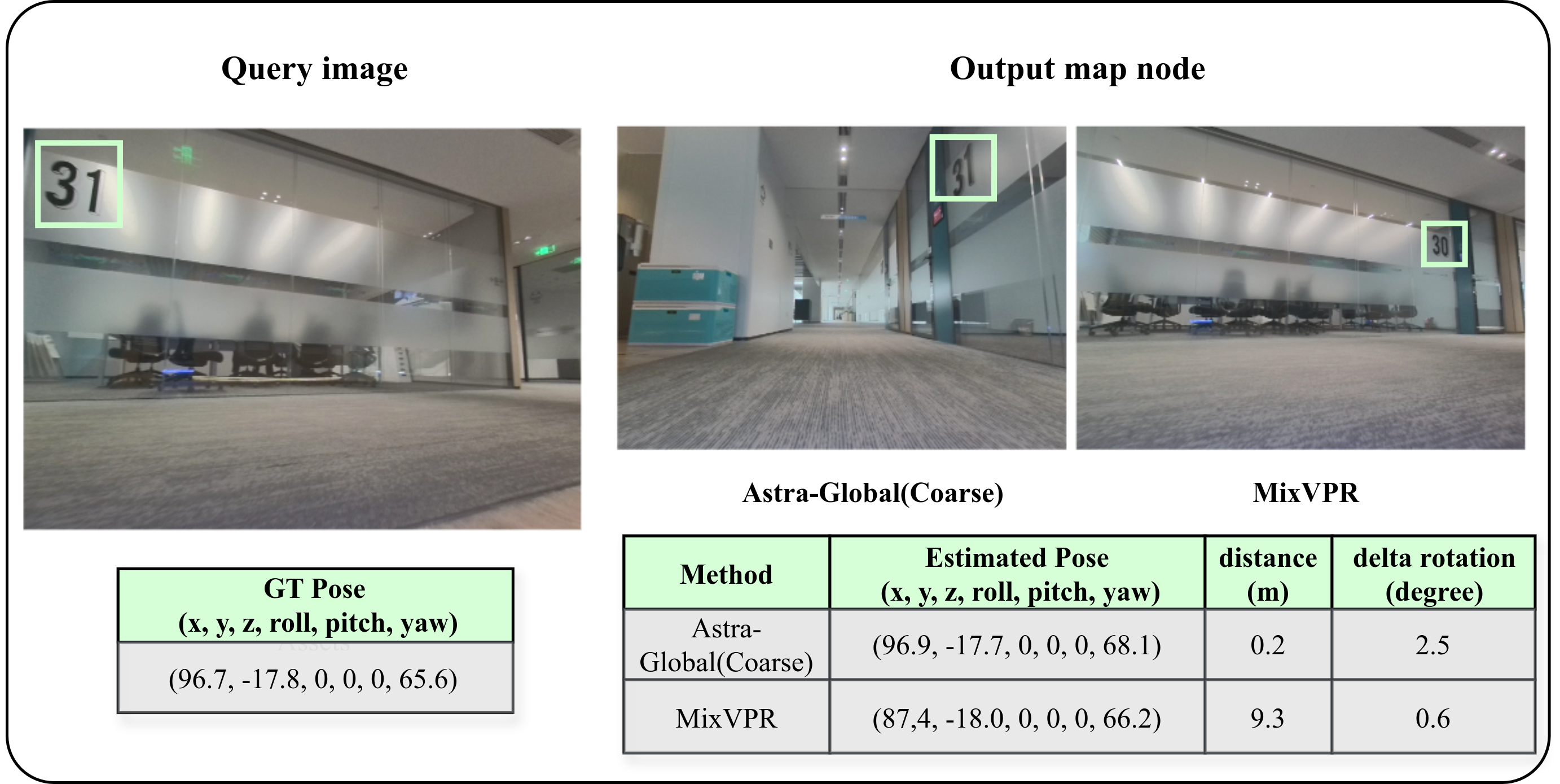}
        \caption{Office Localization}
        \label{fig:compare_vpr_b}
    \end{subfigure}%
    \caption{Astra-Global's robust localization performance in cases where VPR fails} 
    \label{fig:compare_vpr} 
\end{figure}

The results demonstrate that our method significantly outperforms \cite{ali2023mixvpr} in all scenarios. Key advantages include:
\begin{itemize}
\item Better detail capture: VPR, like \cite{ali2023mixvpr}, uses global features, often misses fine details like room numbers. Our method catches these details, avoiding failures in similar scenes with repetitive layouts, as shown in Fig.~\ref{fig:compare_vpr}(a).
\item More robust to viewpoint changes: VPR often has trouble with big viewpoint shifts while \method-Global is more robust as it relies on semantic landmarks. The relative positions between landmarks stay the same even when the camera angle changes. An example is shown in Fig.~\ref{fig:compare_vpr}(b).
\item Higher pose accuracy: In the presence of multiple similar candidate positions, our method leverages landmark spatial relationships to select the best-matching pose, achieving significantly higher accuracy within pose error compared to VPR (as shown in Fig.~\ref{fig:compare_with_vpr}).
\end{itemize}

We also validate the cross-scene generalization capability of \method-Global in a zero-shot transfer experiment: the model is trained on warehouse and office building datasets, then directly deployed to home environments without any parameter fine-tuning. As shown in Fig.~\ref{fig:reloc_example_all}(c) and Fig.~\ref{fig:compare_vpr}, our method achieves 81.8\% pose accuracy under 1m-5° criteria, demonstrating more than 20\% percentage points improvement over MixVPR's 57.7\%. When finetuned on limited home environment data, the performance of \method-Global further boosted to 91.1\%.

\begin{numberedconclusion}
GRPO improves \method-Global’s generalization.
\end{numberedconclusion}

To validate the effectiveness of GRPO in our framework, we conduct comprehensive ablation studies on the coarse localization task, evaluating two training paradigms: (1) supervised fine-tuning (SFT) alone and (2) SFT followed by GRPO reinforcement learning. Tab.~\ref{tab:grpo_ablation} presents the localization accuracy (\%, within 10m and 180°) across different environments. Notably, the Home environment was employed as a zero-shot scenario, enabling an assessment of the generalization capacity of the methods.

\begin{table}[!htbp]
    \centering
    \footnotesize
    \begin{tabular}{lcccc}
        \toprule
        \multirow{2}{*}{Method} & \multirow{2}{*}{\#samples} & \multicolumn{3}{c}{scenario} \\
        \cmidrule(lr){3-5}
        & & Warehouse & Office & Home \\
        \midrule
        SFT-only & 100k        & 89.9 & 93.8 & 93.7 \\
        SFT-only & 300k        & 93.1 & 94.6 & 97.3 \\
        SFT+GRPO & 100k + 20k & 93.3 & 95.3 & 99.9\\
        SFT+GRPO & 100k + 200k & 95.5 & 95.3 & 99.9\\
        \bottomrule
    \end{tabular}
    \caption{Ablation results of GRPO on coarse localization}
    \label{tab:grpo_ablation}
\end{table}
The experimental results demonstrate GRPO's effectiveness in improving performance for coarse localization. In the zero-shot home scene, SFT+GRPO (100k+20k) attains an impressive accuracy of 99.9\%, significantly outperforming SFT-only method (93.7\%) even with a comparable number of samples. Moreover, within the previously-seen warehouse and office environments, GRPO consistently yields performance gains in the range of 0.7-2.4\%. This not only validate its capability to enhance generalization but also highlight its potential to improve sample efficiency through the application of reinforcement learning techniques.

\begin{numberedconclusion}
Flow matching with proposed Masked ESDF loss gives the best local trajectory in \method-Local.
\end{numberedconclusion}

We train our planning head in \method-Local on 10M trajectory samples recorded via human remote control and evaluate the models on two test dataset: ID (in-distribution) and OOD (out-of-distribution).
The OOD dataset exists numerous unseen congested scenarios that are absent from the training data.
We mainly compare our methods with ACT \cite{zhao2023learning} and diffusion policy (DP) \cite{chi2023diffusion} while the pretrained encoders remain the same. We also compare results w/ and w/o the proposed masked ESDF to exam its usefulness.
We use collision rate and velocity as our main evaluation metrics.
As safety and velocity often exhibit a strong trade-off, we also use a score function as a supplementary metric, which is learned from data annotated by human experts.

\begin{figure}[!htbp]
    \centering
    \begin{subfigure}[b]{0.45\textwidth}
        \includegraphics[width=\textwidth]{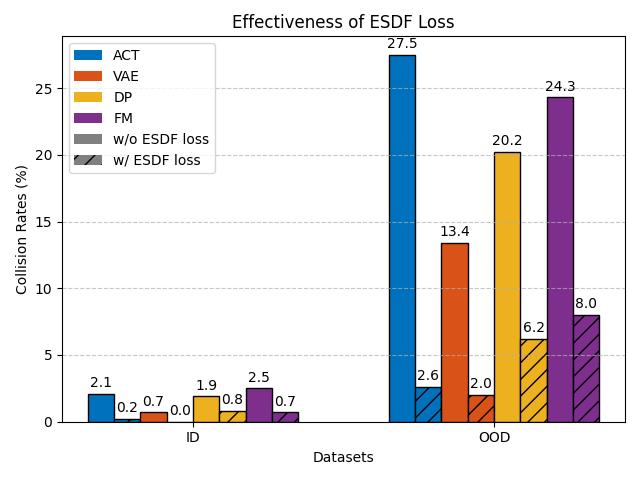}
        \caption{Effectiveness of ESDF loss.}
        \label{exp:esdf_loss}
    \end{subfigure}
    \hfill
    \begin{subfigure}[b]{0.45\textwidth}
        \includegraphics[width=\textwidth]{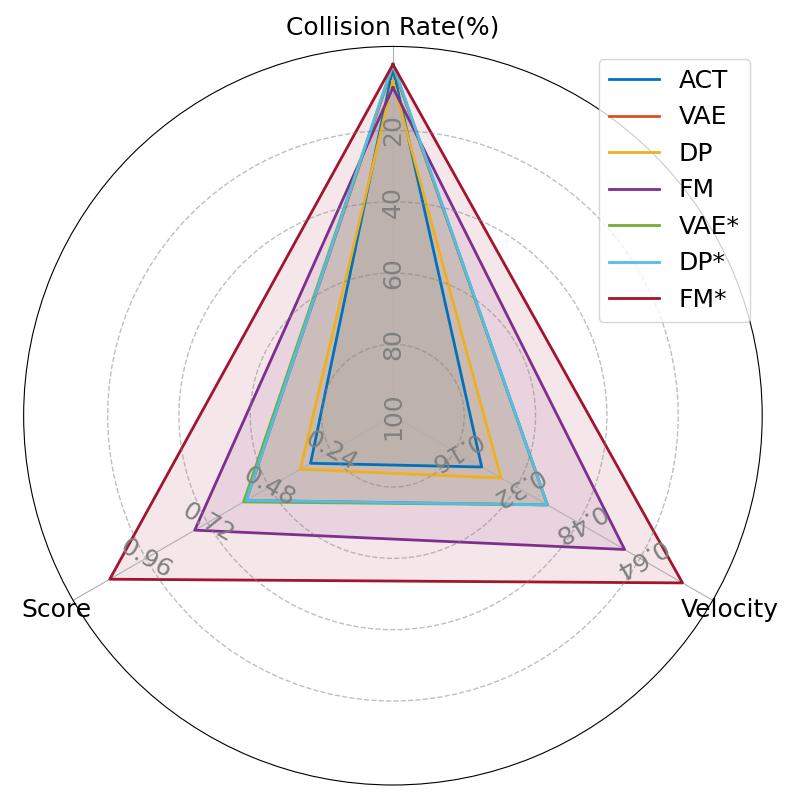}
        \caption{Comparison of planning heads.}
        \label{exp:planning_heads}
    \end{subfigure}
    \caption{Ablation of different planning head \& ESDF loss.}
    \label{exp:esdf_planning_heads}
\end{figure}

The effectiveness of the proposed masked ESDF loss is shown in Fig.~\ref{exp:esdf_loss}.
We can observe that the ESDF loss can significantly reduce the collision rates of all approaches both on ID and OOD datasets.
The comparison of different planning heads on OOD dataset is shown in Fig.~\ref{exp:planning_heads}, where all methods apply the same encoder and ESDF loss. The superscript * represents that we use the trajectory of the highest score out of 10 generated samples.
We can see that FM can achieve higher score and velocity while the collision rate can be maintained on the same level. 
If we sample multiple trajectories, FM* can dominate other approaches in terms of all three metrics.

The results varying different model size $\times$ training dataset size are shown in Tab.~\ref{table:scaling}.
We can see that on the ID dataset, as the model scale and dataset size increases, the collision rate will decrease and the velocity will increase, which aligns with the expectation.
However, on the OOD dataset, the collision rate does not show obvious improvement, while the velocity and score does increase.
Nevertheless, the improvement on the score which reflects the overall performance of trajectories still demonstrate the effectiveness of increasing model and dataset sizes.

\begin{table}[!htbp]
\centering
\begin{tabular}{@{}llcccccc@{}}
\toprule
\multirow{2}{*}{Model Scale} & 
\multirow{2}{*}{Data Scale} & 
\multicolumn{3}{c}{ID} & 
\multicolumn{3}{c}{OOD} \\
\cmidrule(lr){3-5} \cmidrule(lr){6-8}
 & & {Collision rate} & {Velocity} & {Score} & {Collision rate} & {Velocity} & {Score} \\
\midrule
\multirow{3}{*}{Small} 
 & 1M & 3.5\% & 0.63 & 0.96 & 11.4\% & 0.45 & 0.52 \\
 & 5M & 1.4\% & 0.75 & 1.21 & 11.9\% & 0.50 & 0.59 \\
 & 10M & 1.3\% & 0.68 & 1.10 & 11.3\% & 0.47 & 0.54 \\
\addlinespace
\multirow{3}{*}{Medium}
 & 1M & 3.2\% & 0.67 & 0.98 & 12.8\% & 0.42 & 0.48 \\
 & 5M & 1.8\% & 0.78 & 1.15 & 15.7\% & 0.53 & 0.58 \\
 & 10M & 1.1\% & 0.77 & 1.17 & 12.5\% & 0.55 & 0.64 \\
\addlinespace
\multirow{3}{*}{Large}
 & 1M & 1.5\% & 0.80 & 1.26 & 15.3\% & 0.55 & 0.61 \\
 & 5M & 1.0\% & 0.80 & 1.26 & 14.2\% & 0.58 & 0.68 \\
 & 10M & 0.7\% & 0.87 & 1.34 & 8.0\% & 0.60 & 0.77 \\
\bottomrule
\end{tabular}

\smallskip
\footnotesize
\textsuperscript{*} Velocity values are normalized to [0,1]
\caption{Ablation of different model size \& dataset size for Astra-Local planning head.}
\label{table:scaling}
\end{table}

\begin{figure}[!ht]
    \centering
    \begin{subfigure}{\textwidth}
        \includegraphics[width=\textwidth, height=\imgcaseheight]{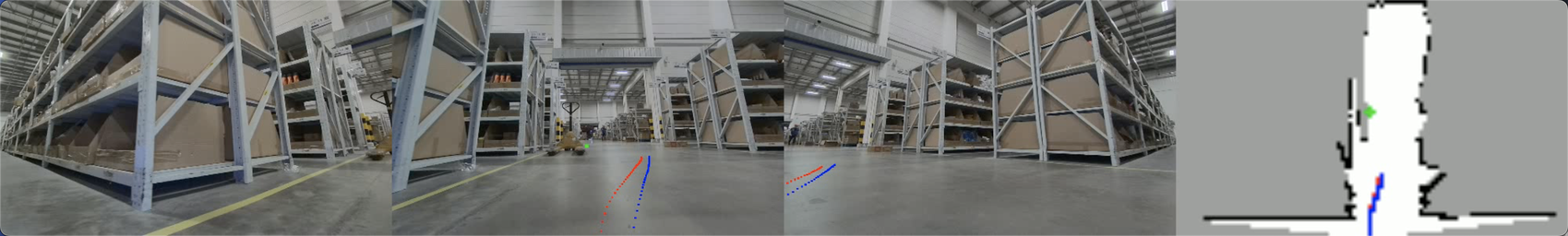} 
    \end{subfigure}
    
    \begin{subfigure}{\textwidth}
        \includegraphics[width=\textwidth, height=\imgcaseheight]{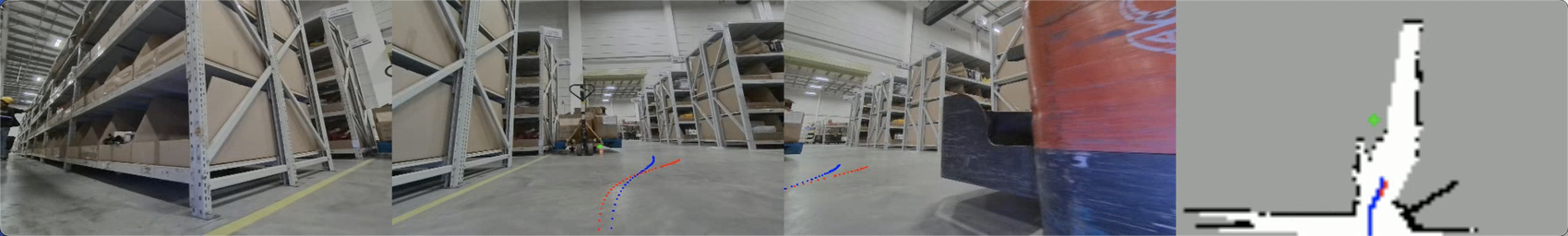}
    \end{subfigure}
    
    \begin{subfigure}{\textwidth}
        \includegraphics[width=\textwidth, height=\imgcaseheight]{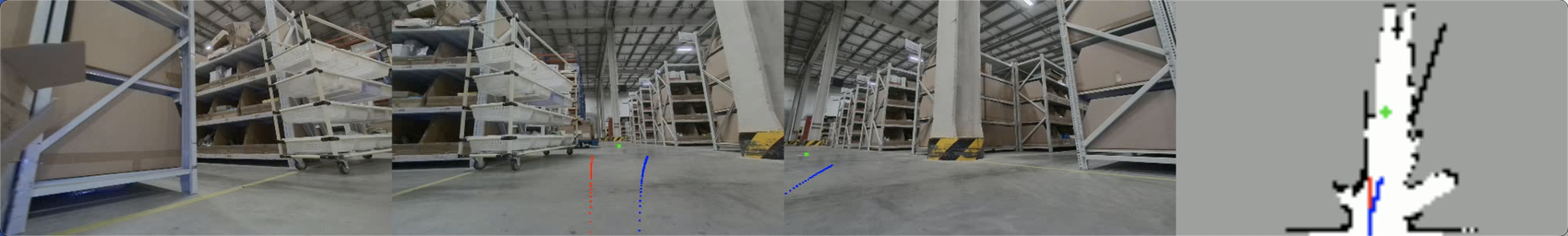}
    \end{subfigure}
    
    \begin{subfigure}{\textwidth}
        \includegraphics[width=\textwidth, height=\imgcaseheight]{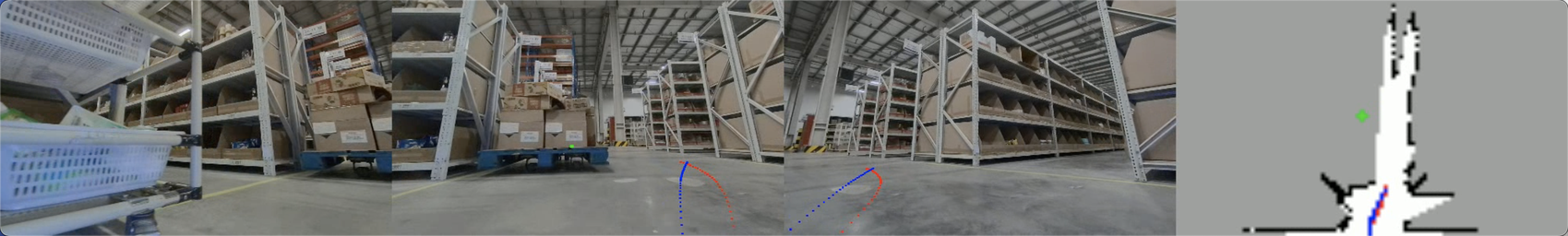}
    \end{subfigure}
    \caption{Planing head case study: interaction with different types of static obstacles.}
    \label{fig:case-static}
\end{figure}

\begin{figure}[!ht]
    \centering
    \begin{subfigure}{\textwidth}
        \includegraphics[width=\textwidth, height=\imgcaseheight]{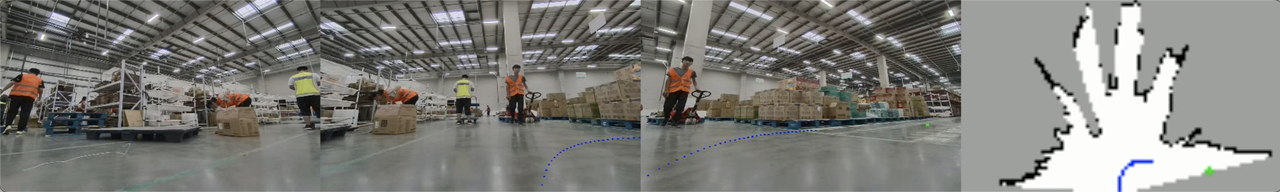} 
    \end{subfigure}
    
    \begin{subfigure}{\textwidth}
        \includegraphics[width=\textwidth, height=\imgcaseheight]{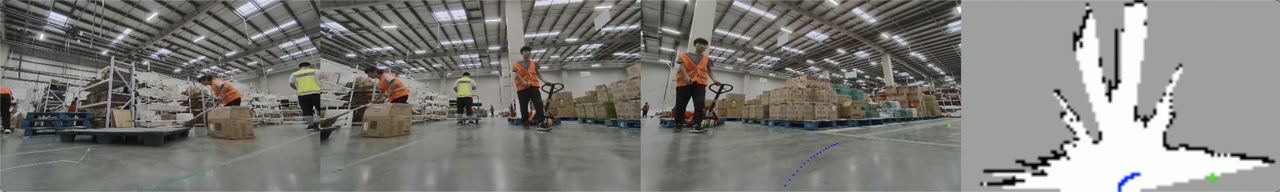}
    \end{subfigure}
    
    \begin{subfigure}{\textwidth}
        \includegraphics[width=\textwidth, height=\imgcaseheight]{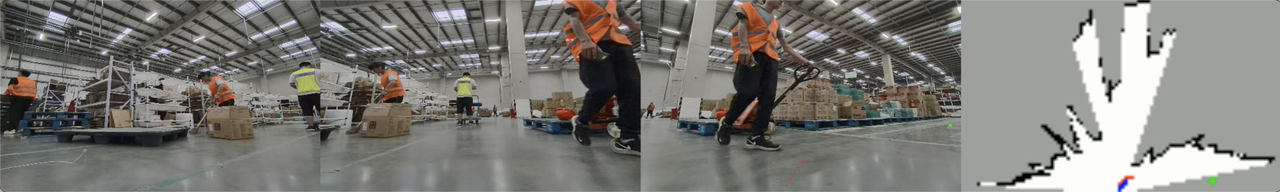}
    \end{subfigure}
    
    \begin{subfigure}{\textwidth}
        \includegraphics[width=\textwidth, height=\imgcaseheight]{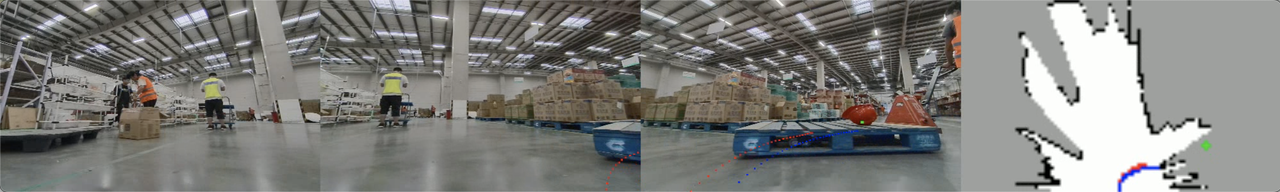}
    \end{subfigure}
    \caption{Planing head case study: interaction with an operator pulling a pallet jack in a warehouse.}
    \label{fig:case-dynamic1}
\end{figure}

\begin{figure}[!ht]
    \centering
    \begin{subfigure}{\textwidth}
        \includegraphics[width=\textwidth, height=\imgcaseheight]{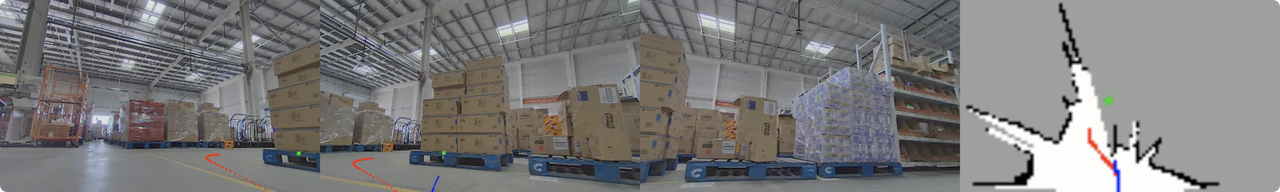} 
    \end{subfigure}
    
    
    
    \begin{subfigure}{\textwidth}
        \includegraphics[width=\textwidth, height=\imgcaseheight]{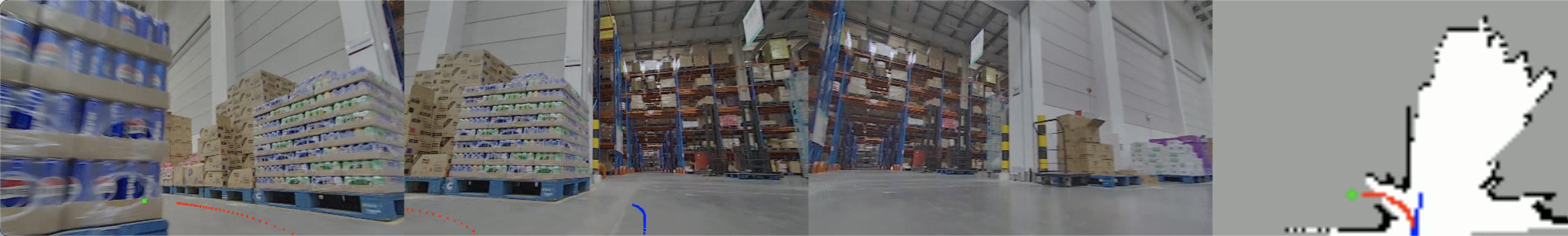}
    \end{subfigure}
    \caption{Planing head case study: corner cases}
    \label{fig:case-bad}
\end{figure}

In addition, we present open-loop qualitative results in Fig.~\ref{fig:case-static},~\ref{fig:case-dynamic1}, and~\ref{fig:case-bad}. The left three images depict the tri-camera color views, and the rightmost one shows the 2D occupancy map. The red trajectories indicate the ground truth, while the blue ones are the model's outputs, and the green squares mark the local goal.
Fig.~\ref{fig:case-static} depicts scenarios where the robots interact with various static obstacles such as picking carts, forklifts, and pallet jacks. Our model can successfully plan a collision-free path to navigate around these obstacles, as shown in the figure.
Fig.~\ref{fig:case-dynamic1} illustrates the interaction between the robot and an operator moving with a pallet jack. When the operator is at a distance, the predicted trajectory tends to turn right in front of them. However, when the operator gets closer, the trajectory shortens rapidly, indicating deceleration. As the operator is about to leave, the trajectory lengthens again, corresponding to an acceleration.

However, there are also some corner cases. Representative examples are shown in Fig.~\ref{fig:case-bad}.
In the first case, while the trajectory remains collision-free, it fails to circumvent the obstacle, demonstrating the limitations of relying solely on collision rate. In the second case, the model commits a directional error, resulting in choosing an incorrect path. 

\begin{numberedconclusion}
Proposed transformer encoder in \method-Local is effective for odometry estimation by fusing multi-sensor multi-frame inputs.
\end{numberedconclusion}

We evaluate the odometry head in \method-Local on our multi-modal dataset containing synchronized image sequences, IMU and wheel measurements, together with ground truth poses. We trained our model and our re-implementation of \cite{bevodom} which serve as our baseline on this dataset. 

As in \cite{bevodom}, it processes consecutive image frames by first extracting Bird's Eye View (BEV) features through a depth-aware perspective-to-BEV encoder. A correlation volume is computed from adjacent BEV features, followed by MLP-based regression to estimate relative poses (3-DoF). Consistent with the original methodology, we adopted a 7×7 search window configuration with a feature resolution of 0.2 m/pixel. This implementation achieved mean positional error of 0.006m and angular error of 0.0014 rad per frame. Subsequent trajectory propagation using these relative poses demonstrated approximately 5\% relative error compared to ground truth trajectories.  

In contrast, our transformer-based odometry head employs temporal modeling of multi-frame sensor data. Specifically, the network architecture integrates both measurements from current frame and historical sensor inputs from the preceding 9 frames to predict inter-frame motion. Note that our BEV correlation tokenizer uses identical configuration parameters as our re-implementation of \cite{bevodom}. We compare different variant of the odometry head in \method-Local to the baseline \cite{bevodom}. Additionally, to exam the effect of the temporal modeling and multi-sensor fusion, we trained \method-Local (bev-only) where we only fuse multi-frame visual data in our transformer encoder, \method-Local (BEV + IMU) where wheel data was not used  and \method-Local where all sensor inputs were used.

\begin{figure}[!ht]
    \centering
    \begin{subfigure}[b]{0.32\textwidth}
        \includegraphics[width=\textwidth]{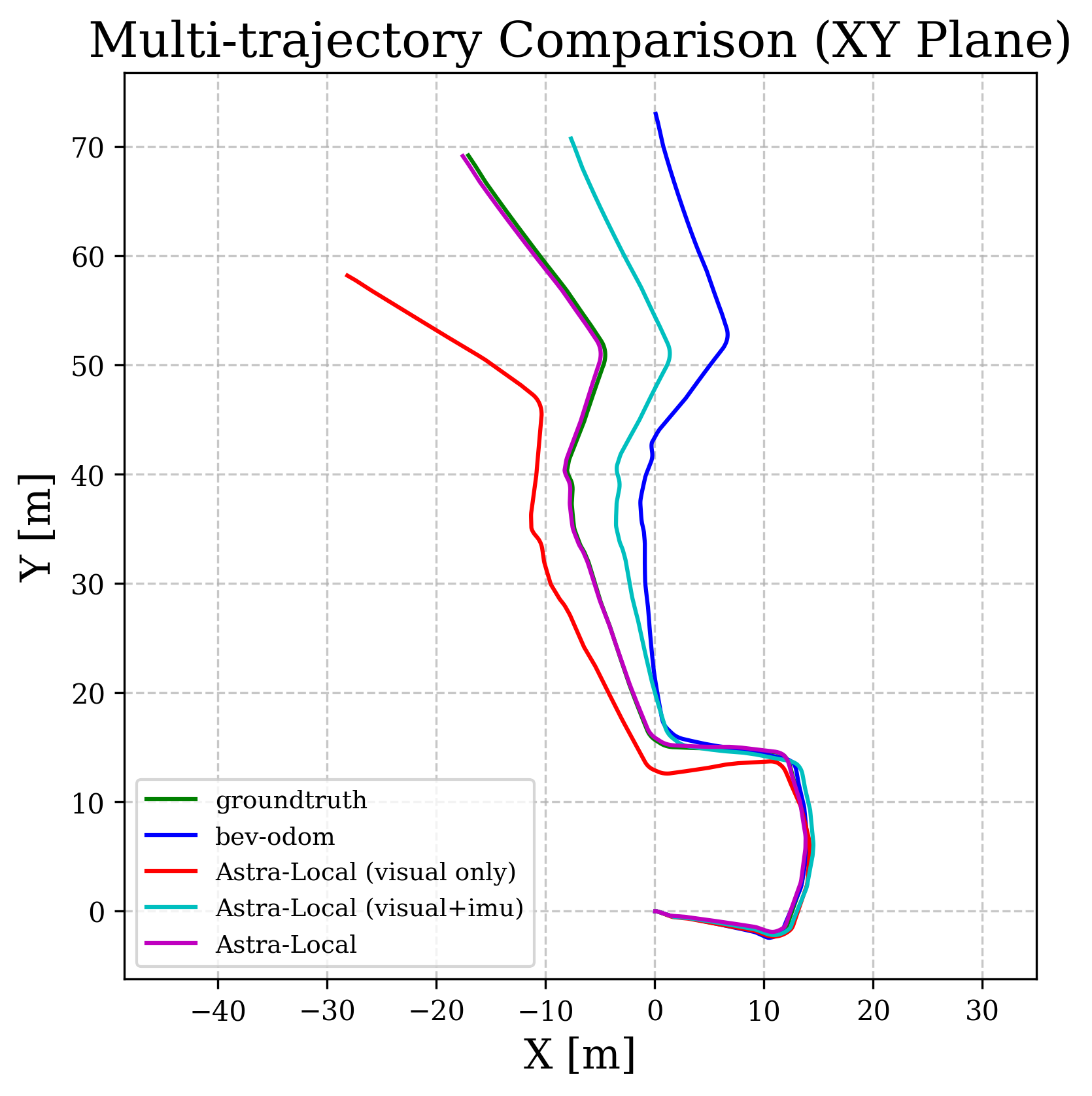}
        \caption{Sequence 1}
        \label{fig:sub1}
    \end{subfigure}
    \hfill
    \begin{subfigure}[b]{0.32\textwidth}
        \includegraphics[width=\textwidth]{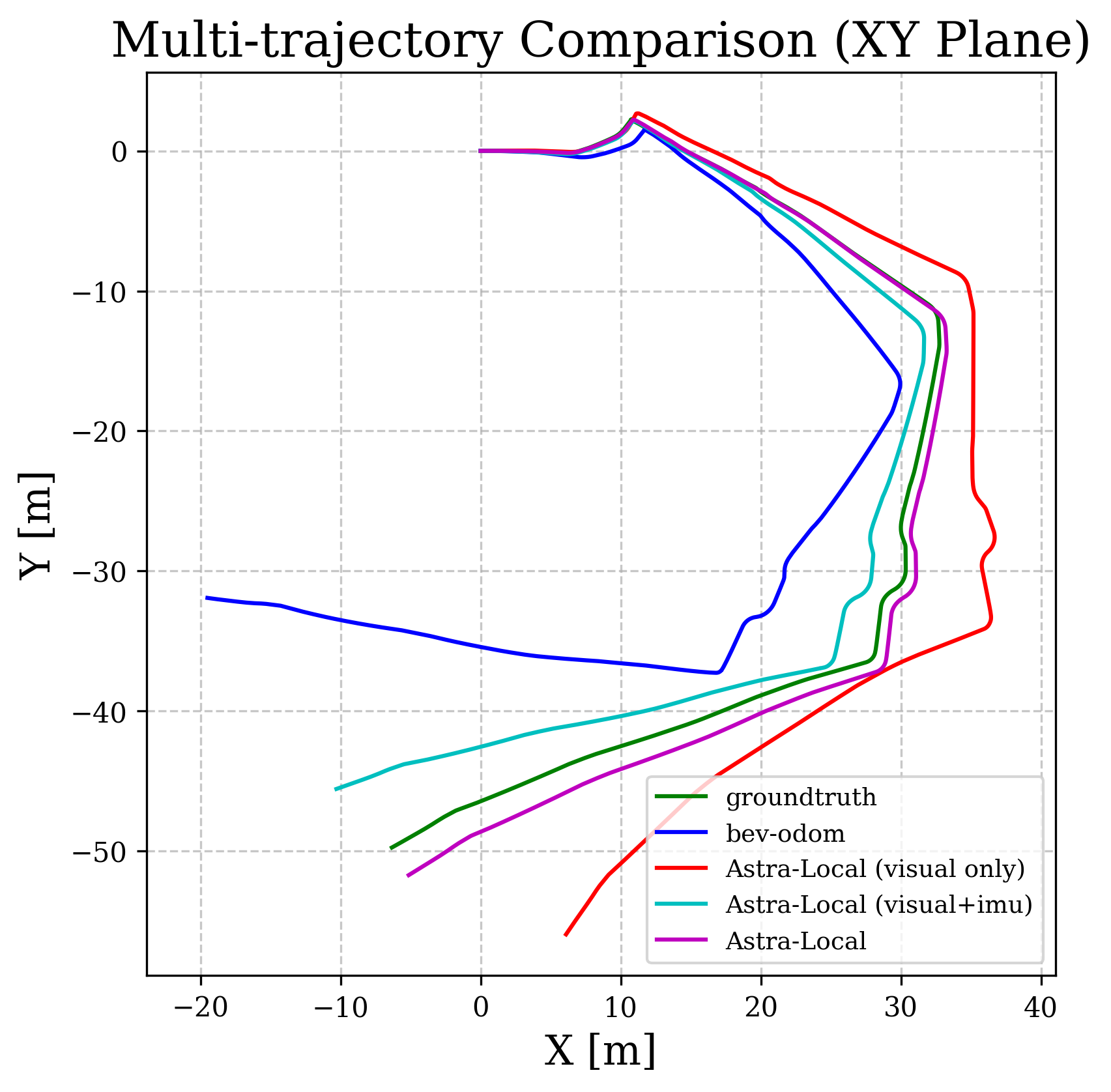}
        \caption{Sequence 2}
        \label{fig:sub2}
    \end{subfigure}
    \hfill
    \begin{subfigure}[b]{0.32\textwidth}
        \includegraphics[width=\textwidth]{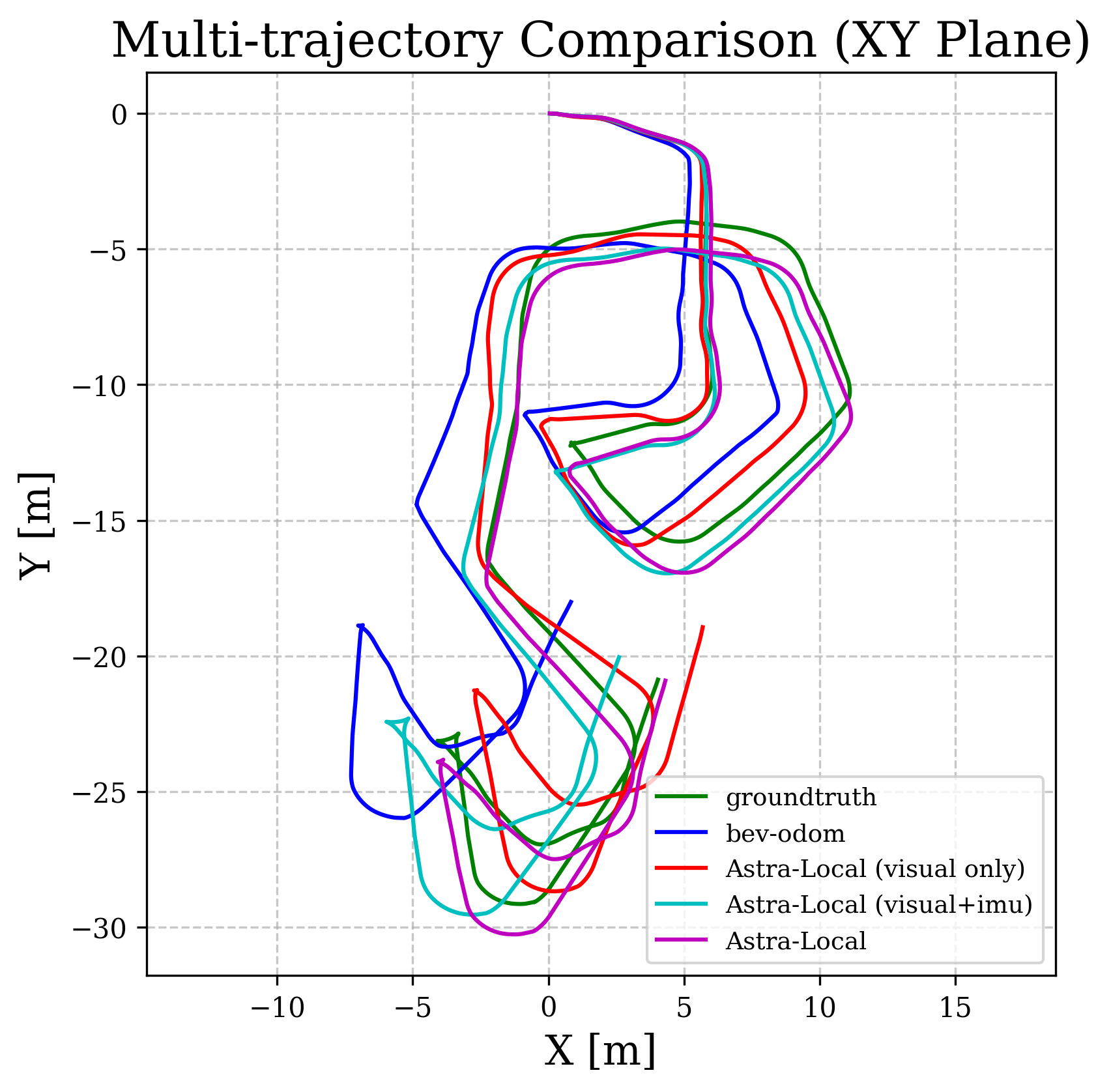}
        \caption{Sequence 3}
        \label{fig:sub3}
    \end{subfigure}
    \caption{Odometry trajectory comparisons between \method-Local and baselines. }
    \label{fig:odom_trajectory}
\end{figure}

Tab.~\ref{table:odometry} and Fig.~\ref{fig:odom_trajectory} present quantitative results and trajectory visualizations, showing significant improvements over the two-frame BEV-ODOM baseline. Specifically, incorporating IMU measurements boosts rotational estimation accuracy, cutting the overall trajectory error to around 2\%. Moreover, integrating wheel data further enhances scale stability and estimation accuracy. These enhancements highlight the advantages of combining temporal fusion and multi-sensor data in our odometry head.

\begin{table}[!ht]
    \centering
    \footnotesize
    \begin{tabular}{cccc}
    \toprule
         Methods&  RTE(\%)&  RRE($^{\circ}$/10m)&  ATE(m)\\
         \midrule
         BEV-ODOM&  5.46\%&  6.36&  1.27\\
         Astra-Local (visual only)&  3.13\%&  2.85&  1.08\\
 Astra-Local (visual + imu)& 2.04\%& 1.19&0.48\\
         Astra-Local &  1.92\%&  0.66&  0.26\\
         \bottomrule
    \end{tabular}
\caption{Comparison of odometry performance}
\label{tab:my_label}
\label{table:odometry}
\end{table}

\begin{numberedconclusion}
The 4D Spatial-Temporal Encoder in \method-Local provides robust feature for downstream tasks.
\end{numberedconclusion}

We examine the effectiveness of the 4D Spatial-Temporal encoder in \method-Local by comparing two downstream tasks: the occupancy prediction and the local path planning. 

Tab.~\ref{tab:pretrain_ablation} shows the results of occupancy prediction where we took the 3D spatial encoder and train it on an occupancy prediction dataset w/ and w/o the pretrained weights. The self-supervised pretraining of the 3D encoder improves the results by a decent margin.

\begin{table}[h!tbp]
\centering
\footnotesize
\begin{tabular}{lc*{3}{c}}
\toprule
\multirow{2}{*}{Pretrained Weights} & \multicolumn{2}{c}{Office Building } & \multicolumn{2}{c}{Warehouse } \\
\cmidrule(lr){2-3} \cmidrule(lr){4-5} 
& Total IoU &  Obstacle IoU & Total IoU &  Obstacle IoU \\
\midrule
 w/o pretrain & 41.10 & 30.78 & 40.91 & 22.45 \\
 w/ pretrain & 42.16 & 32.26 & 42.55 & 25.81 \\
\bottomrule
\end{tabular}
\caption{Effect of 3D spatial encoder on downstream occupancy prediction task.}
\label{tab:pretrain_ablation}
\end{table}

Tab.~\ref{tab:encoderplanning} reveals that, compared to using only a 3D spatial encoder, the addition of the 4D Spatial-Temporal module can increase the overall velocity and decrease the collision rate in most settings. This demonstrates the significance of prediction in local path planning.

\begin{table}[h!tbp]
\centering
\footnotesize
\begin{tabular}{lccccc}
\toprule
\multirow{2}{*}{Encoder} & \multicolumn{2}{c}{ID} & \multicolumn{2}{c}{OOD} \\
\cmidrule(lr){2-3} \cmidrule(lr){4-5}
 & Collision rate & Velocity & Collision rate & Velocity \\
\midrule
\textbf{VAE + ESDF} & & & & \\
\quad 3D encoder & 0.6\% & 0.75 & 8.4\% & 0.35 \\
\quad 4D encoder & 0.0\% & 0.72 & 2.0\% & 0.40 \\
\addlinespace
\textbf{FM + ESDF} & & & & \\
\quad 3D encoder & 0.9\% & 0.78 & 10.3\% & 0.52 \\
\quad 4D encoder & 0.7\% & 0.87 & 8.0\% & 0.60 \\
\bottomrule
\end{tabular}
\caption{Planning Performance Comparison of Different Encoders}
\label{tab:encoderplanning}
\end{table}


We also investigate the effect of model scale and dataset size for the 4D Spatial-Temporal Encoder. The full pretraining dataset contains about 10M training samples. We progressively varied the scale of pretraining dataset from 250k to 1.5M and finally to 10M samples. We also evaluate by finetuning the pretrained models on occupancy prediction tasks. It can be seen in Tab.~\ref{tab:encoder_size} that the model performs the best at the scale of 1.5M. When it comes to the 10M, the performance degrades a little. The observed performance plateau could potentially be attributed to the model's limited capacity, constraining its ability to benefit from larger-scale pretraining dataset. Therefore, we also experimented with larger backbones, switching from the original ViT-S to ViT-L architectures, while keeping the scale of pretraining dataset at 10M. We can see that with increased model capacity, the largest model now performs the best.


\begin{table}[h!tbp]
\centering
\footnotesize
\begin{tabular}{c *{2}{cc}}
\toprule
\multirow{2}{*}{Dataset \& Model Scale } & 
\multicolumn{2}{c}{Office Building} & 
\multicolumn{2}{c}{Warehouse} \\
\cmidrule(lr){2-3} \cmidrule(lr){4-5}
 & Total IoU & Obstacle IoU  & Total IoU & Obstacle IoU \\
\midrule
250K (ViT-S) & 40.14 & 30.84 & 36.64 & 25.68 \\
1.5M (ViT-S) & 42.30 & 33.17 & 37.57 & 27.12 \\
10M (ViT-S) & 41.65 & 32.40 & 36.27 & 25.37 \\
10M (ViT-L) & 43.27 & 33.94 & 41.64 & 26.91 \\
\bottomrule
\end{tabular}
\caption{Ablation of dataset and model size of 3D encoder for downstream occupancy prediction.}
\label{tab:encoder_size}
\end{table}



\begin{figure}[h!tbp]
  \centering
  \begin{minipage}[b]{1.0\textwidth}
  \begin{subfigure}[b]{0.16\textwidth}
    \includegraphics[width=\linewidth, height=0.65\linewidth]{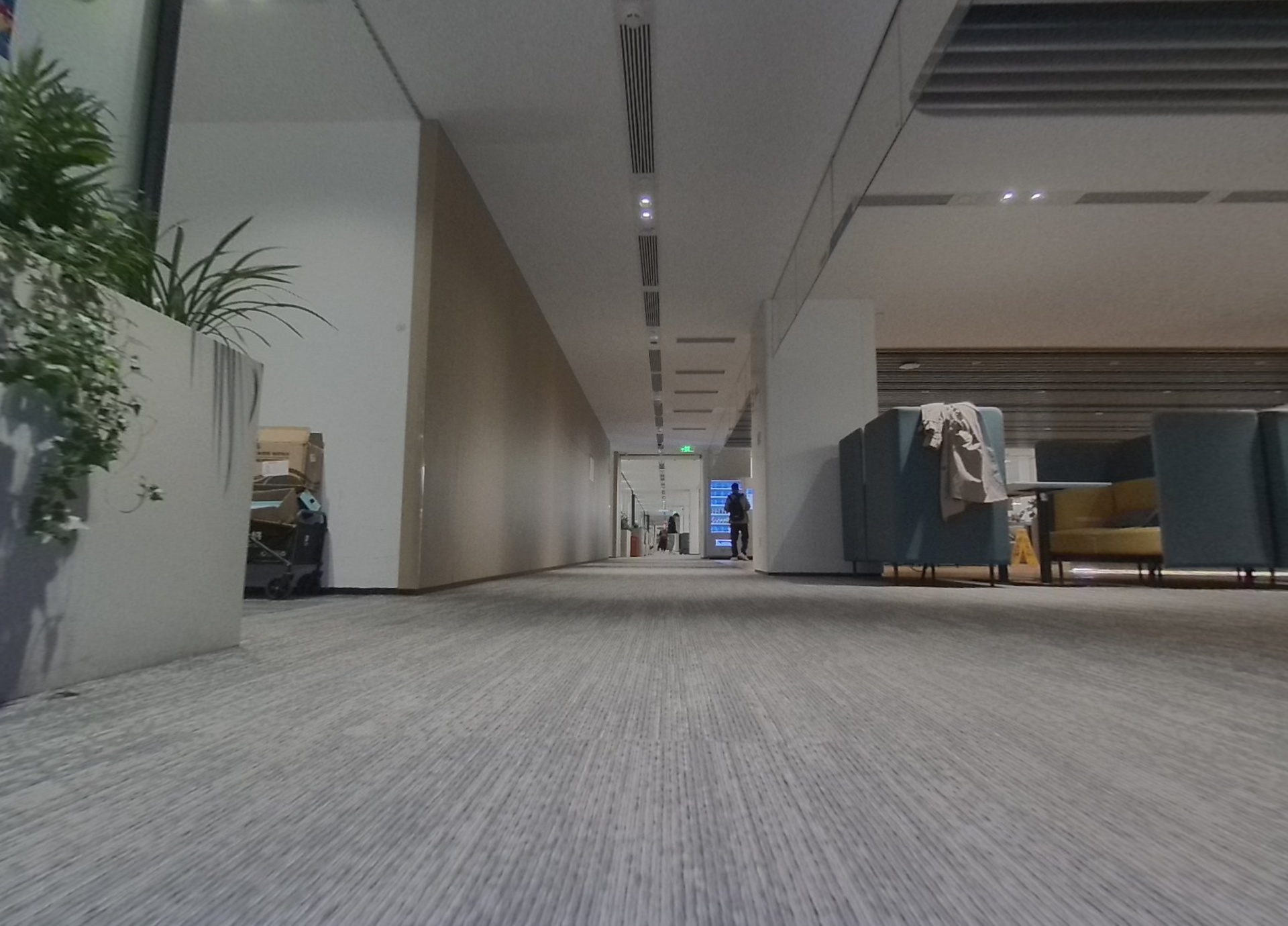}
  \end{subfigure}%
  \hspace{0.005\textwidth}%
  \begin{subfigure}[b]{0.16\textwidth}
    \includegraphics[width=\linewidth, height=0.65\linewidth]{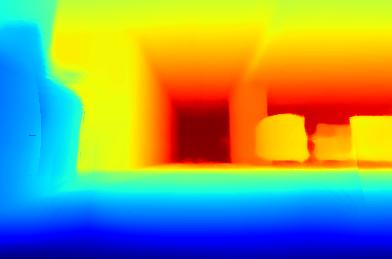}
  \end{subfigure}%
  \hspace{0.005\textwidth}%
  \begin{subfigure}[b]{0.16\textwidth}
    \includegraphics[width=\linewidth, height=0.65\linewidth]{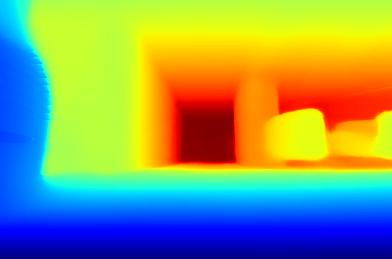}
  \end{subfigure}%
  \hspace{0.005\textwidth}%
  \begin{subfigure}[b]{0.16\textwidth}
    \includegraphics[width=\linewidth, height=0.65\linewidth]{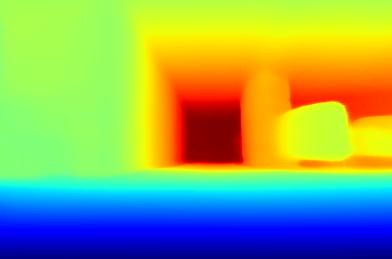}
  \end{subfigure}%
  \hspace{0.005\textwidth}%
    \begin{subfigure}[b]{0.16\textwidth}
    \includegraphics[width=\linewidth, height=0.65\linewidth]{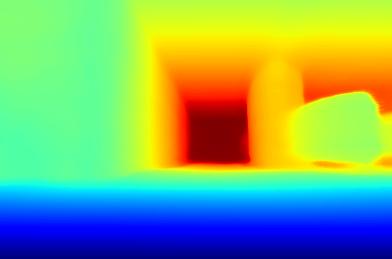}
  \end{subfigure}%
  \hspace{0.005\textwidth}%
    \begin{subfigure}[b]{0.16\textwidth}
    \includegraphics[width=\linewidth, height=0.65\linewidth]{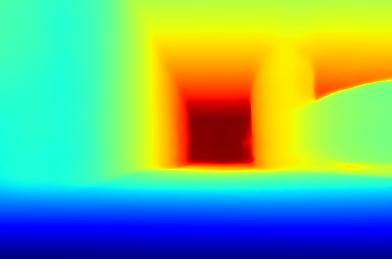}
  \end{subfigure}%
  \end{minipage}

  \begin{minipage}[b]{1.0\textwidth}
  \begin{subfigure}[b]{0.16\textwidth}
    \includegraphics[width=\linewidth, height=0.65\linewidth]{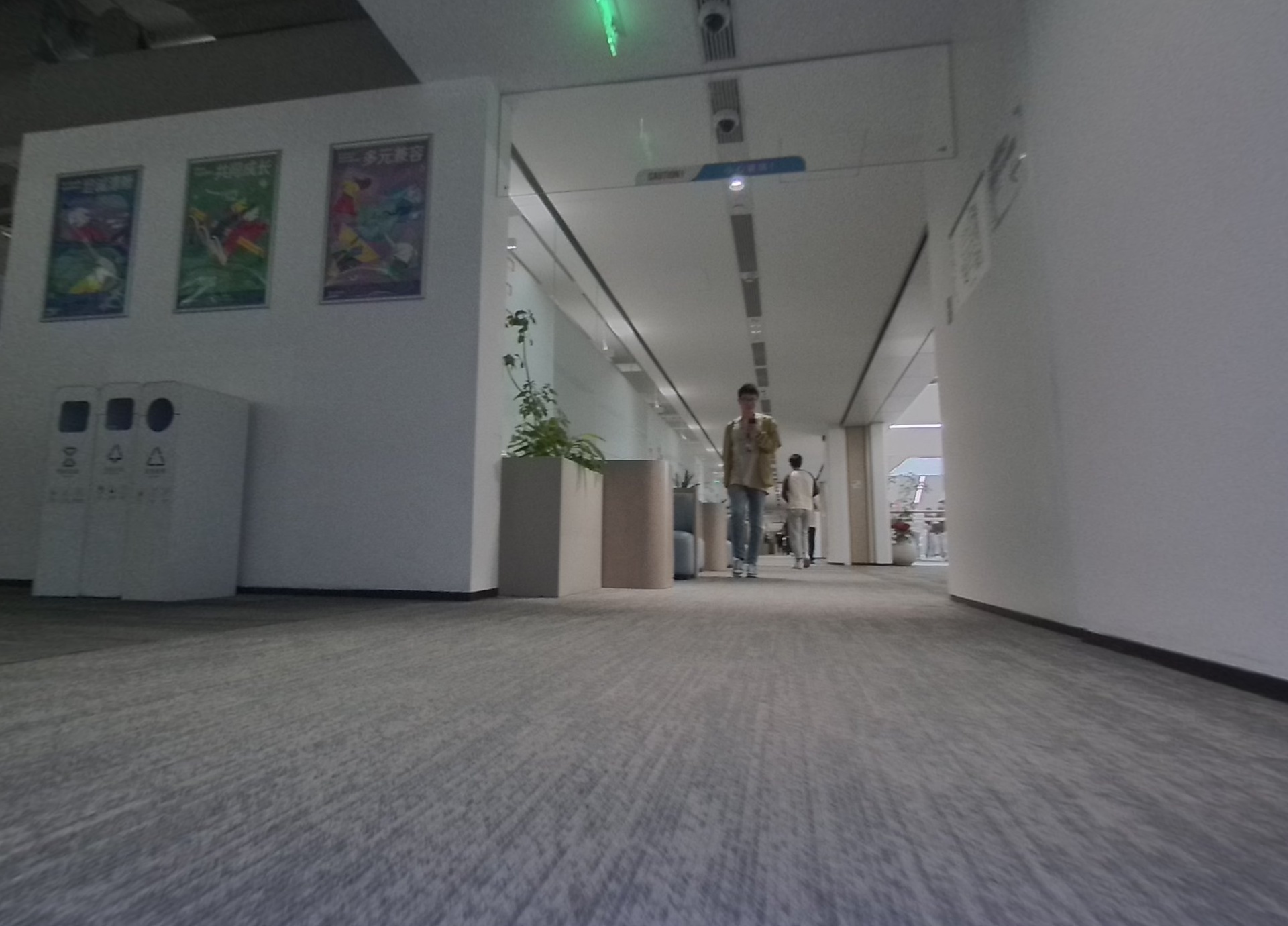}
  \end{subfigure}%
  \hspace{0.005\textwidth}%
  \begin{subfigure}[b]{0.16\textwidth}
    \includegraphics[width=\linewidth, height=0.65\linewidth]{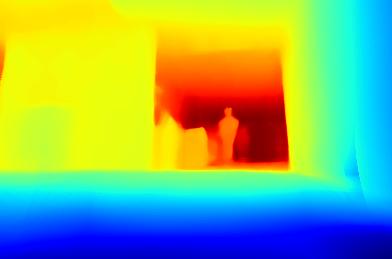}
  \end{subfigure}%
  \hspace{0.005\textwidth}%
  \begin{subfigure}[b]{0.16\textwidth}
    \includegraphics[width=\linewidth, height=0.65\linewidth]{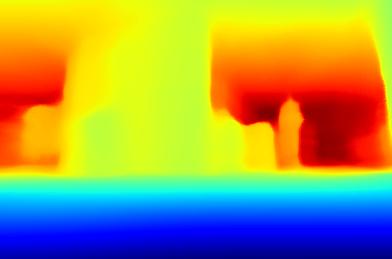}
  \end{subfigure}%
  \hspace{0.005\textwidth}%
  \begin{subfigure}[b]{0.16\textwidth}
    \includegraphics[width=\linewidth, height=0.65\linewidth]{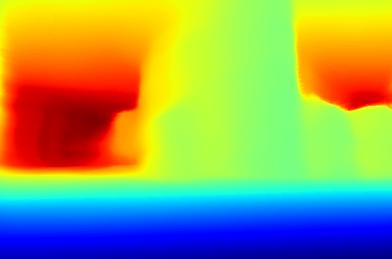}
  \end{subfigure}%
  \hspace{0.005\textwidth}%
    \begin{subfigure}[b]{0.16\textwidth}
    \includegraphics[width=\linewidth, height=0.65\linewidth]{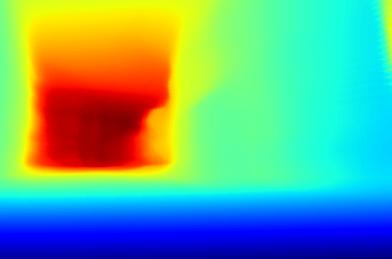}
  \end{subfigure}%
  \hspace{0.005\textwidth}%
    \begin{subfigure}[b]{0.16\textwidth}
    \includegraphics[width=\linewidth, height=0.65\linewidth]{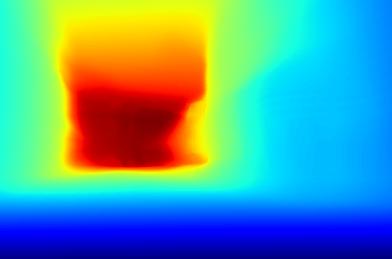}
  \end{subfigure}%
  \end{minipage}

  \begin{minipage}[b]{1.0\textwidth}
  \begin{subfigure}[b]{0.16\textwidth}
    \includegraphics[width=\linewidth, height=0.65\linewidth]{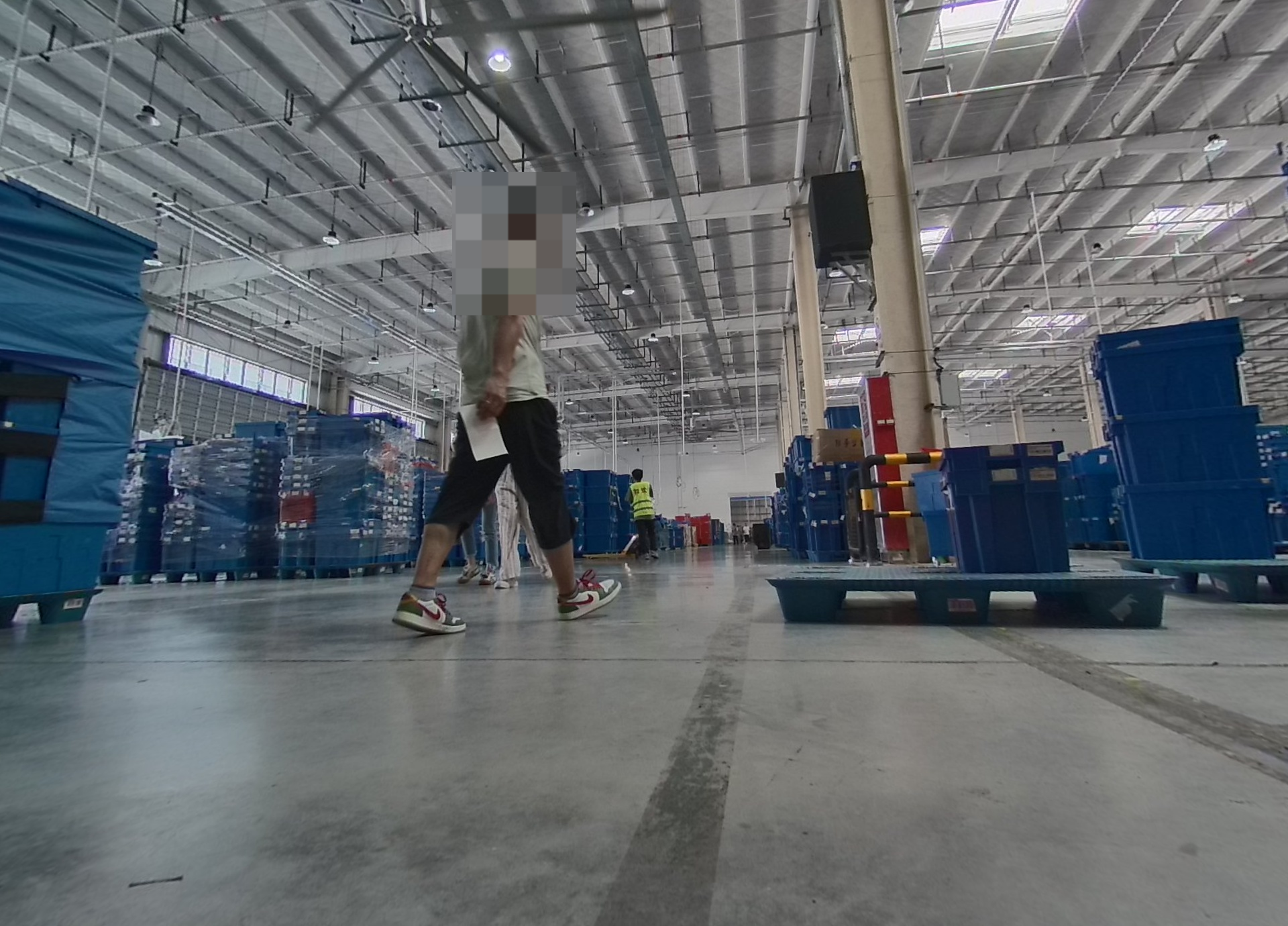}
  \end{subfigure}%
  \hspace{0.005\textwidth}%
  \begin{subfigure}[b]{0.16\textwidth}
    \includegraphics[width=\linewidth, height=0.65\linewidth]{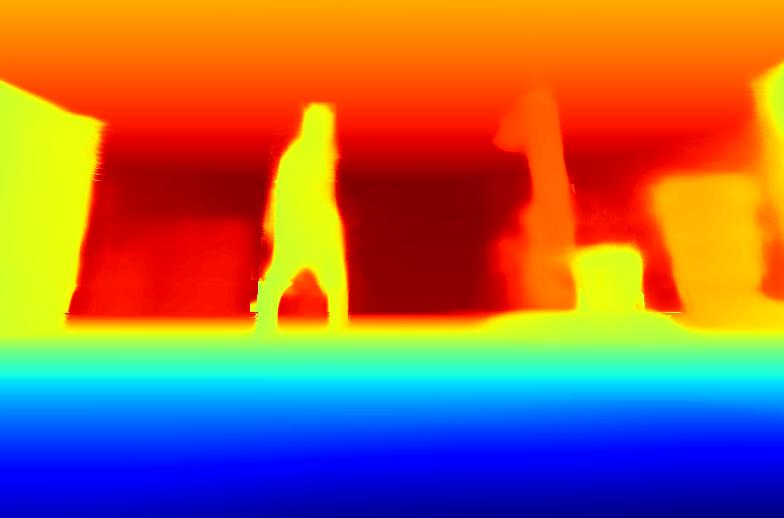}
  \end{subfigure}%
  \hspace{0.005\textwidth}%
  \begin{subfigure}[b]{0.16\textwidth}
    \includegraphics[width=\linewidth, height=0.65\linewidth]{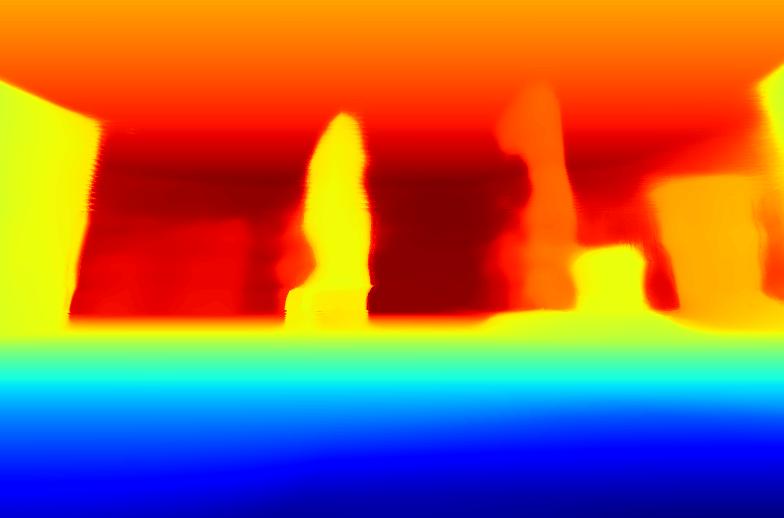}
  \end{subfigure}%
  \hspace{0.005\textwidth}%
  \begin{subfigure}[b]{0.16\textwidth}
    \includegraphics[width=\linewidth, height=0.65\linewidth]{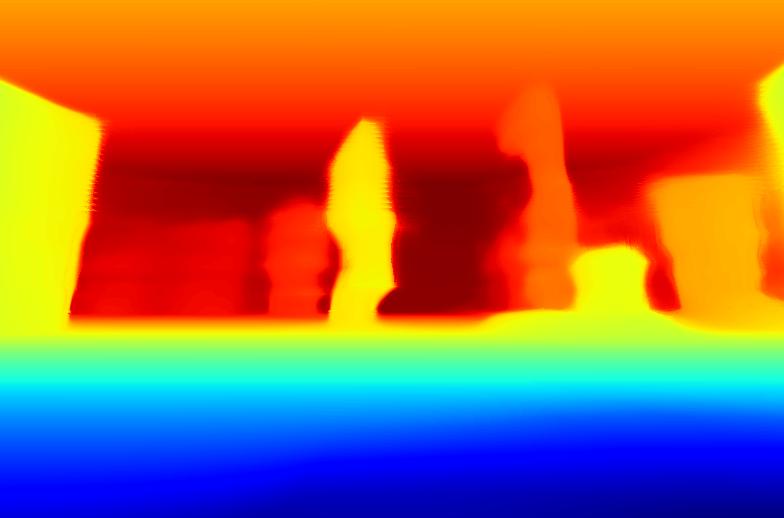}
  \end{subfigure}%
  \hspace{0.005\textwidth}%
    \begin{subfigure}[b]{0.16\textwidth}
    \includegraphics[width=\linewidth, height=0.65\linewidth]{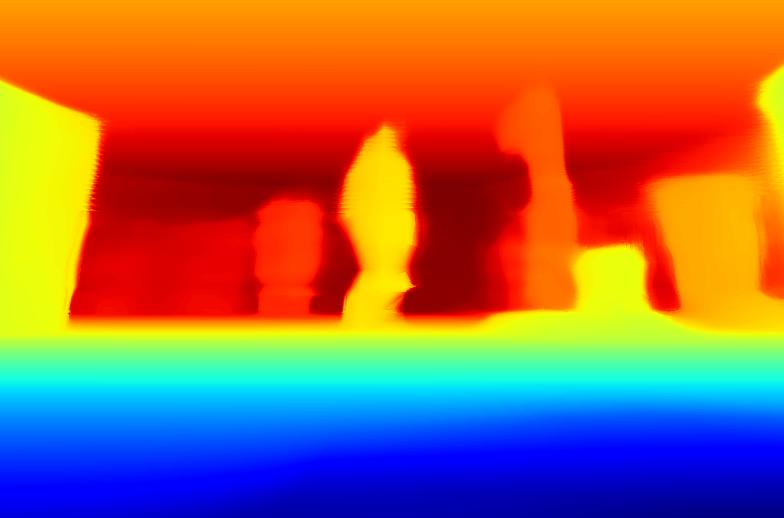}
  \end{subfigure}%
  \hspace{0.005\textwidth}%
    \begin{subfigure}[b]{0.16\textwidth}
    \includegraphics[width=\linewidth, height=0.65\linewidth]{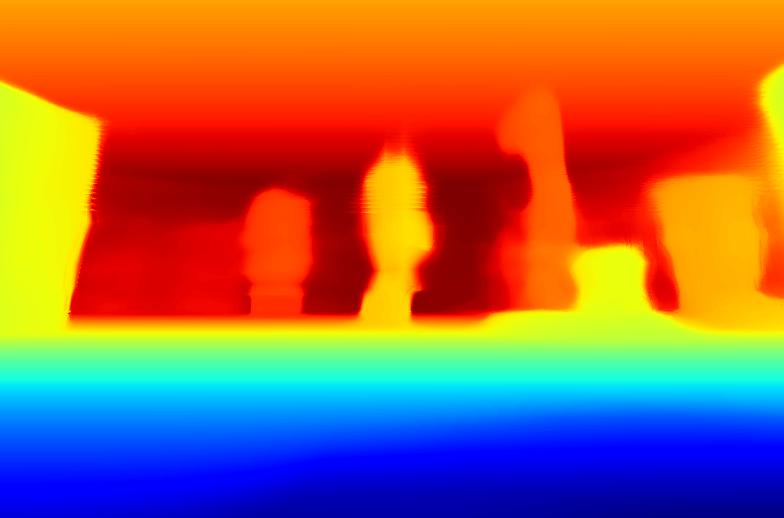}
  \end{subfigure}%
  \end{minipage}

  \begin{minipage}[b]{1.0\textwidth}
  \begin{subfigure}[b]{0.16\textwidth}
    \includegraphics[width=\linewidth, height=0.65\linewidth]{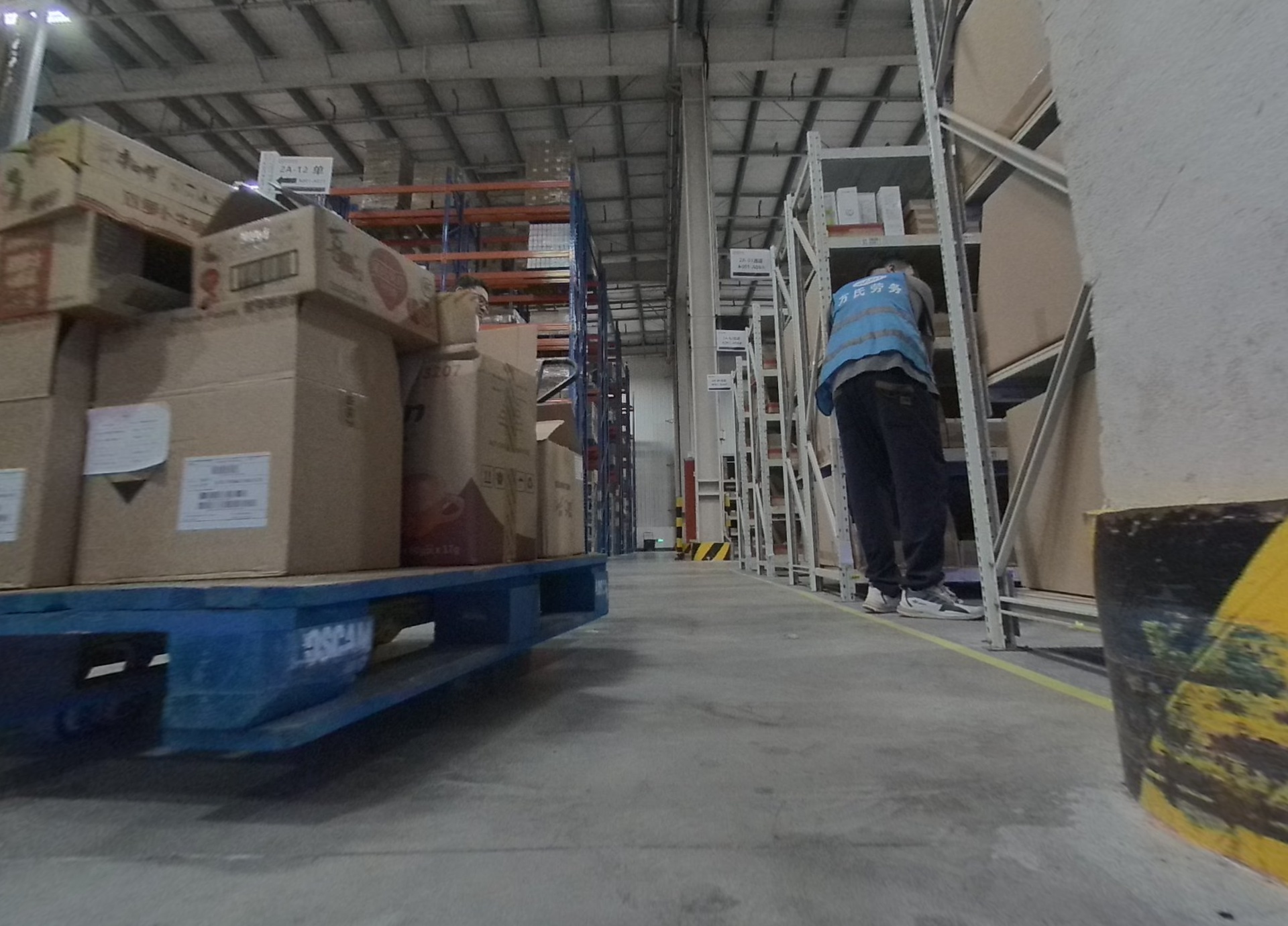}
    \vspace{-1 em}
  \end{subfigure}%
  \hspace{0.005\textwidth}%
  \begin{subfigure}[b]{0.16\textwidth}
    \includegraphics[width=\linewidth, height=0.65\linewidth]{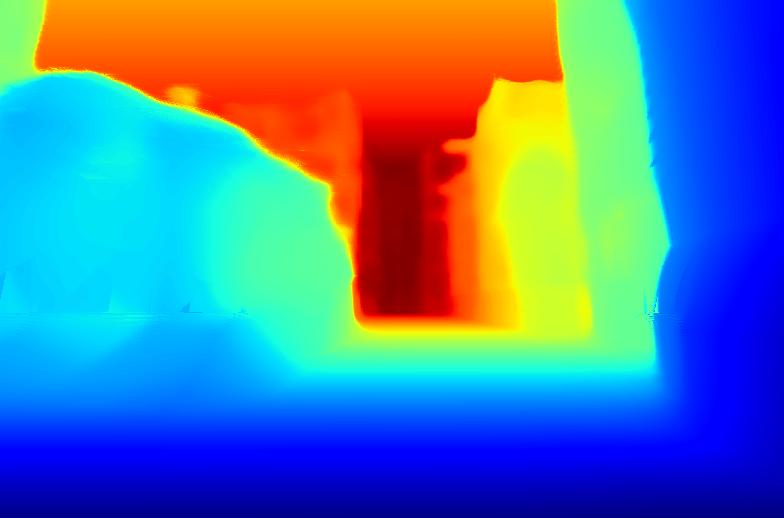}
    \vspace{-1 em}
  \end{subfigure}%
  \hspace{0.005\textwidth}%
  \begin{subfigure}[b]{0.16\textwidth}
    \includegraphics[width=\linewidth, height=0.65\linewidth]{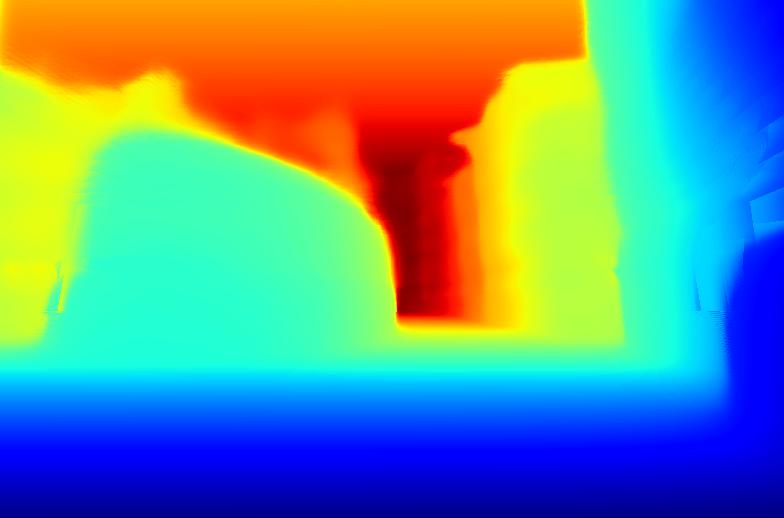}
    \vspace{-1 em}
  \end{subfigure}%
  \hspace{0.005\textwidth}%
  \begin{subfigure}[b]{0.16\textwidth}
    \includegraphics[width=\linewidth, height=0.65\linewidth]{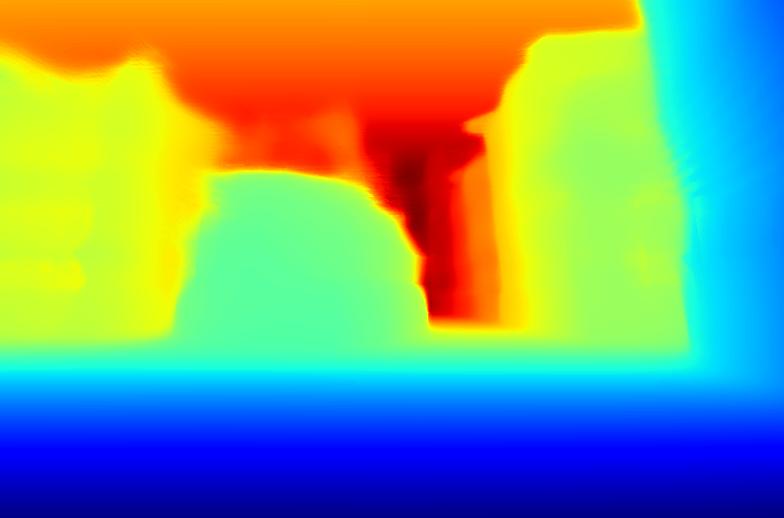}
    \vspace{-1 em}
  \end{subfigure}%
    \hspace{0.005\textwidth}%
  \begin{subfigure}[b]{0.16\textwidth}
    \includegraphics[width=\linewidth, height=0.65\linewidth]{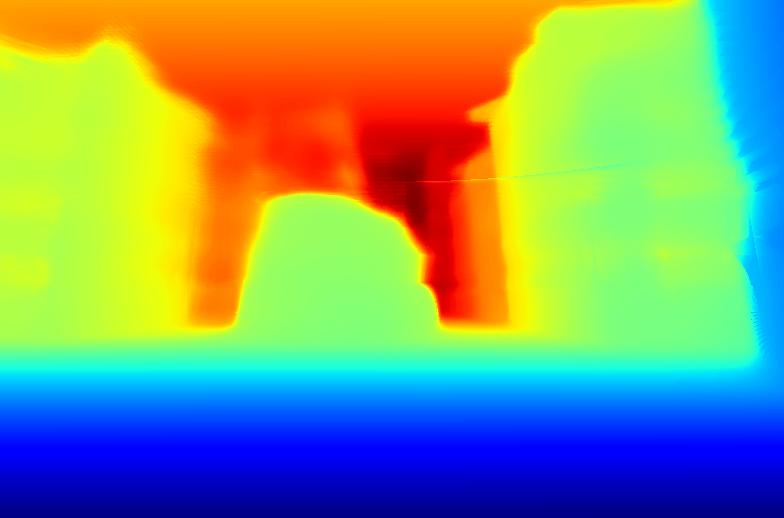}
    \vspace{-1 em}
  \end{subfigure}%
    \hspace{0.005\textwidth}%
  \begin{subfigure}[b]{0.16\textwidth}
    \includegraphics[width=\linewidth, height=0.65\linewidth]{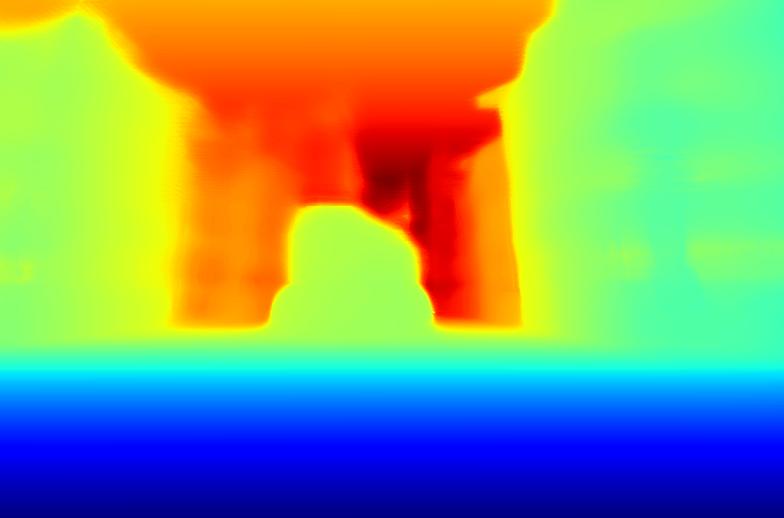}
    \vspace{-1 em}
  \end{subfigure}%
  \end{minipage}
\caption{Visualization for 4D Spatial-Temporal Encoder. First column shows the current color image. Second column is the rendered depth from voxel features in current frame. Column 3-6 shows rendered depth from predicted voxel features in 0.5s, 1s, 1.5s, 2s. Colder colors indicate objects that are closer, while warmer colors indicate objects that are farther away.}
\label{fig:voxel_predict}
\end{figure}

Finally, we compare the 4D encoder with the SOTA occupancy forecasting methods. 
The 4D encoder takes the past 4 frames of voxel features as input and output the future 4 frames, where each frame of voxel features is produced by the pretrained 3D encoder. We compare our 4D encode with the our re-implementation of OccWorld \cite{zheng2024occworld} and Cam4DOcc \cite{ma2024cam4docc} which are trained using Equation \ref{eq:4dloss}, with the 3D encoder frozen. We evaluate the predicted voxel features from different aspects. As the voxel features can be converted into SDF via the fixed MLP from 3D encoder which can further be transformed to occupancy grid. We evaluate the predicted occupancy grid. Besides, the voxel features can be rendered into depth and color images, we use AbsRel and PSNR to evaluate the quality of the rendered future depth and color image respectively. The results are shown in Tab.~\ref{tab:benchmark}. Benefiting from the multi-scale architecture and the DiT block, voxel features predicted from our 4D encoder can generate better occupancy, depth and color images. In Fig.~\ref{fig:voxel_predict}, we show some examples of rendered depth from the predicted future voxels. More visualization results can be found on our website.


\begin{table}[h!tbp]
\centering
\footnotesize
\begin{tabular}{lc*{7}{c}}
\toprule
\multirow{2}{*}{Methods} & \multicolumn{4}{c}{Warehouse}  \\
\cmidrule(lr){2-5}
& Total IoU & Obstacle IoU & AbsRel (Depth) & PSNR (Color) \\
\midrule
OccWorld  & 82.90 & 41.05 & 0.1752 & 29.1943  \\
Cam4DOcc  & 81.17 & 51.01 & 0.1376 & 28.9919  \\
Astra-Local  & 84.80 & 56.66 & 0.1179 & 29.3021  \\
\bottomrule
\end{tabular}
\caption{Comparison of 4D encoder to the SOTA methods}
\label{tab:benchmark}
\end{table}

\vspace{20pt}

\section{Conclusion and Future Work}
\label{conclusion}
In this report, we introduce our newly developed \method, a dual-model architecture for mobile robot navigation. Unlike traditional modular systems, it integrates key navigation tasks—such as goal/robot localization and local path planning—into two cohesive models. \method-Global, a multimodal LLM, localizes multimodal queries (e.g., text or images) within a pre-built hybrid topological-semantic map. Our results demonstrate its robustness across diverse environments and ability to generalize to unseen scenarios with minimal or no additional data.
\method-Local, a multi-sensor multi-task network focused on path planning and odometry estimation, leverages a pretrained 4D spatial-temporal encoder to generate robust features for downstream tasks. Its planning head uses a novel masked ESDF loss combined with flow matching to minimize collision rates, while the odometry head fuses multi-sensor data via a transformer encoder for accurate relative pose estimation, effectively integrating diverse sensor inputs.

In the future, we plan to deploy \method across broader scenarios and continue enhancing its robustness and generalization capabilities. For \method-Global: Although the current map representation balances information loss and token length, it may still lack certain semantic details critical for localization. We intend to investigate alternative map compression approaches to retain essential semantic information while optimizing efficiency. Additionally, the current localization relies solely on single-frame observations, which can fail in scenarios where even humans would struggle (e.g., featureless or highly repetitive environments). To address this, we plan to enable the robot to actively explore its surroundings and incorporate temporal reasoning into the model, leveraging sequential observations for more robust localization. For \method-Local: Our real-robot deployments reveal a non-negligible fallback rate, stemming from both the model’s generalization limitations and the tendency of the rule-based fallback system to trigger erroneously in edge cases. We aim to enhance the model’s robustness to out-of-distribution (OOD) scenarios and redesign the fallback system to be a more seamless, integrated component of the system. Furthermore, we plan to integrate instruction-following capabilities into the model, enabling natural human-robot interaction and expanding usability in dynamic, human-centric environments.
\newpage
\section{Contributions and Acknowledgments}
The names are sorted in alphabetical order of the last name.

\subsection*{\textcolor{brown}{Core Contributors}}
Sheng Chen, Peiyu He, Jiaxin Hu, Ziyang Liu, Yansheng Wang, Tao Xu, Chi Zhang, Chongchong Zhang


\subsection*{Contributors}
Chao An, Shiyu Cai, Duo Cao, Kangping Chen, Shuai Chu, Tianwei Chu, Mingdi Dan, Min Du, Weiwei Fang, Pengyou Fu, Junkai Hu, Xiaowei Jiang, Zhaodi Jiang, Fuxuan Li, Jun Li, Minghui Li, Mingyao Li, Yanchang Li, Zhibin Li, Guangming Liu, Kairui Liu, Lihao Liu, Weizhi Liu, Xiaoshun Liu, Yufei Liu, Yunfei Liu, Qiang Lu, Yuanfei Luo, Xiang Lv, Hongying Ma, Sai Ma, Lingxian Mi, Sha Sa, Hongxiang Shu, Lei Tian, Chengzhi Wang, Jiayu Wang, Kaijie Wang, Qingyi Wang, Renwen Wang, Tao Wang, Wei Wang, Xirui Wang, Chao Wei, Xuguang Wei, Zijun Xia, Zhaohao Xiao, Tingshuai Yan, Liyan Yang, Yifan Yang, Zhikai Yang, Zhong Yin, Li Yuan, Liuchun Yuan, Chi Zhang, Jinyang Zhang, Junhui Zhang, Linge Zhang, Zhenyi Zhang, Zheyu Zhang, Dongjie Zhu

\subsection*{Team Lead}
Hang Li, Yangang Zhang

\clearpage

\bibliographystyle{unsrt}
\bibliography{main}


\end{document}